\documentclass{article}

\usepackage[nonatbib,preprint]{neurips_2024}
\usepackage[numbers]{natbib}

\usepackage[utf8]{inputenc} 
\usepackage[T1]{fontenc}    
\usepackage{hyperref}       
\usepackage{url}            
\usepackage{booktabs}       
\usepackage{amsfonts}       
\usepackage{nicefrac}       
\usepackage{microtype}      
\usepackage{xcolor}         

\usepackage{amsmath}
\usepackage{amssymb}
\usepackage{mathtools}
\usepackage{amsthm}

\usepackage[capitalize,noabbrev]{cleveref}

\theoremstyle{plain}
\newtheorem{theorem}{Theorem}[section]

\theoremstyle{definition}
\newtheorem{definition}[theorem]{Definition}

\theoremstyle{remark}

\usepackage{xspace}
\newcommand{\name}{Standardized Embedder\xspace}
\usepackage[bb=dsserif]{mathalpha}
\usepackage{multirow}
\usepackage{multicol}
\usepackage{comment}
\usepackage{enumitem}
\setlist[itemize]{align=parleft,left=3pt}
\usepackage{graphicx}
\usepackage{subfigure}
\usepackage{amsmath}
\usepackage{amssymb}
\usepackage{mathtools}
\usepackage{amsthm}
\usepackage{float}
\usepackage{soul}

\usepackage{amsmath,amsfonts,bm}

\def\eqref#1{equation~\ref{#1}}

\def\1{\bm{1}}

\DeclareMathAlphabet{\mathsfit}{\encodingdefault}{\sfdefault}{m}{sl}
\SetMathAlphabet{\mathsfit}{bold}{\encodingdefault}{\sfdefault}{bx}{n}

\newcommand{\E}{\mathbb{E}}

\title{Towards Fundamentally Scalable Model Selection: \\Asymptotically Fast Update and Selection}

\author{%
  Wenxiao Wang\thanks{Work done during Wenxiao’s internship at Sony AI.}\\
  University of Maryland\\
  \texttt{wwx@umd.edu} \\
  \And
  Weiming Zhuang\\
  Sony AI\\
  \texttt{weiming.zhuang@sony.com}\\
  \And
  Lingjuan Lyu\thanks{Corresponding Author.}\\
  Sony AI\\
  \texttt{lingjuan.lv@sony.com}
}

\begin{document}

\maketitle

\begin{abstract}
The advancement of deep learning technologies is bringing new models every day, motivating the study of scalable model selection. An ideal model selection scheme should minimally support two operations efficiently over a large pool of candidate models: \textbf{update}, which involves either adding a new candidate model or removing an existing candidate model, and \textbf{selection}, which involves locating highly performing models for a given task. However, previous solutions to model selection require high computational complexity for at least one of these two operations.
In this work, we target \ul{fundamentally (more) scalable model selection that supports asymptotically fast update and asymptotically fast selection at the same time}.
\textbf{Firstly}, we define \textbf{isolated model embedding}, a family of model selection schemes supporting asymptotically fast update and selection: With respect to the number of candidate models $m$, the update complexity is O(1) and the selection consists of a single sweep over $m$ vectors in addition to O(1) model operations.
Isolated model embedding also implies several desirable properties for applications.
\textbf{Secondly}, we present \textbf{Standardized Embedder}, an empirical realization of isolated model embedding. We assess its effectiveness by using it to select representations from a pool of 100 pre-trained vision models for classification tasks and measuring the performance gaps between the selected models and the best candidates with a linear probing protocol. Experiments suggest our realization is effective in selecting models with competitive performances and highlight isolated model embedding as a promising direction towards model selection that is fundamentally (more) scalable. 
\end{abstract}

\section{Introduction}

\begin{figure}[tbp!]
\begin{center}
\subfigure[Brute-force]{
\includegraphics[width=0.31\linewidth]{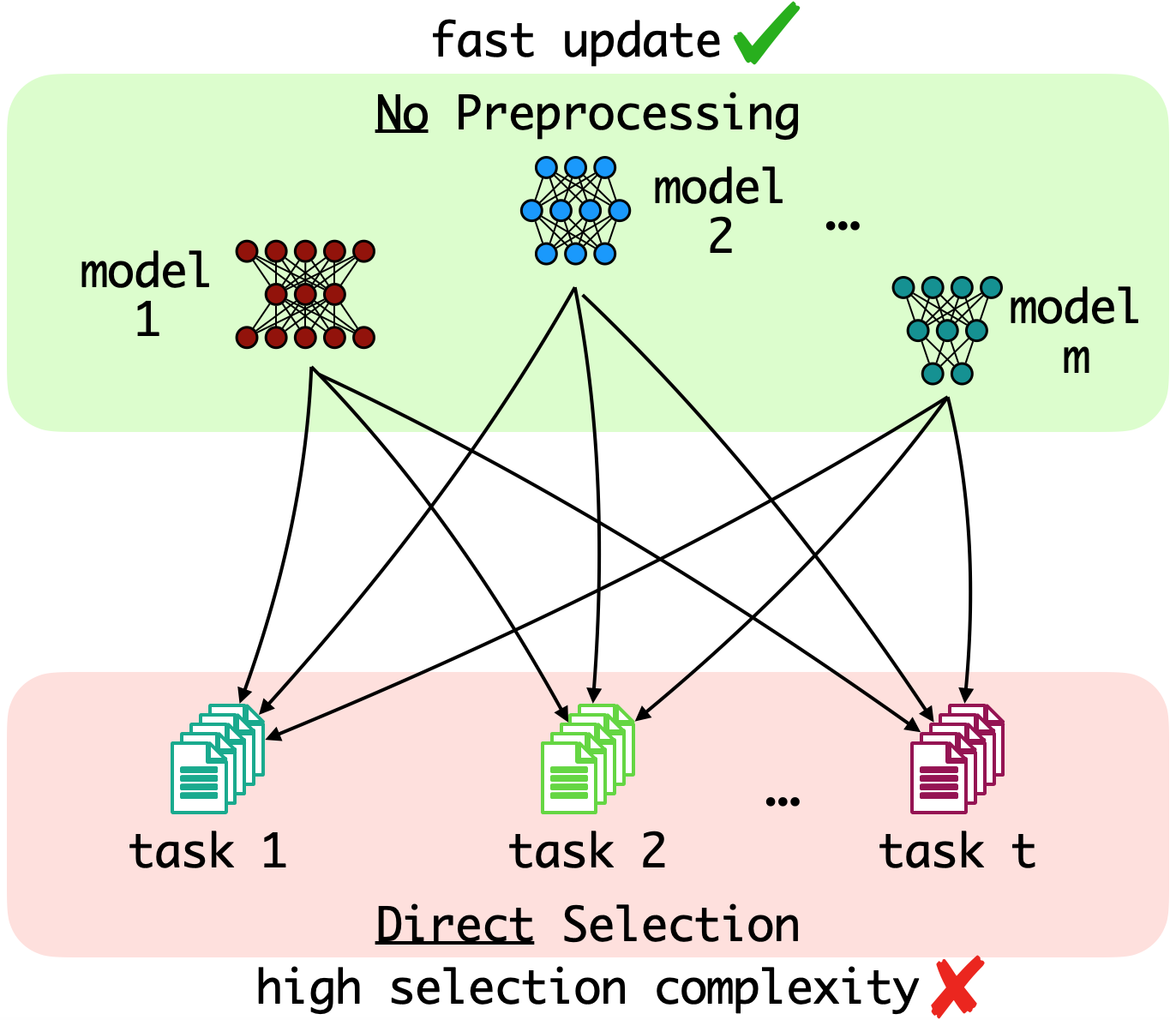}
\label{fig:PrepFree_Model_Selection}
}
\hfill
\subfigure[Model Embedding]{
\includegraphics[width=0.31\linewidth]{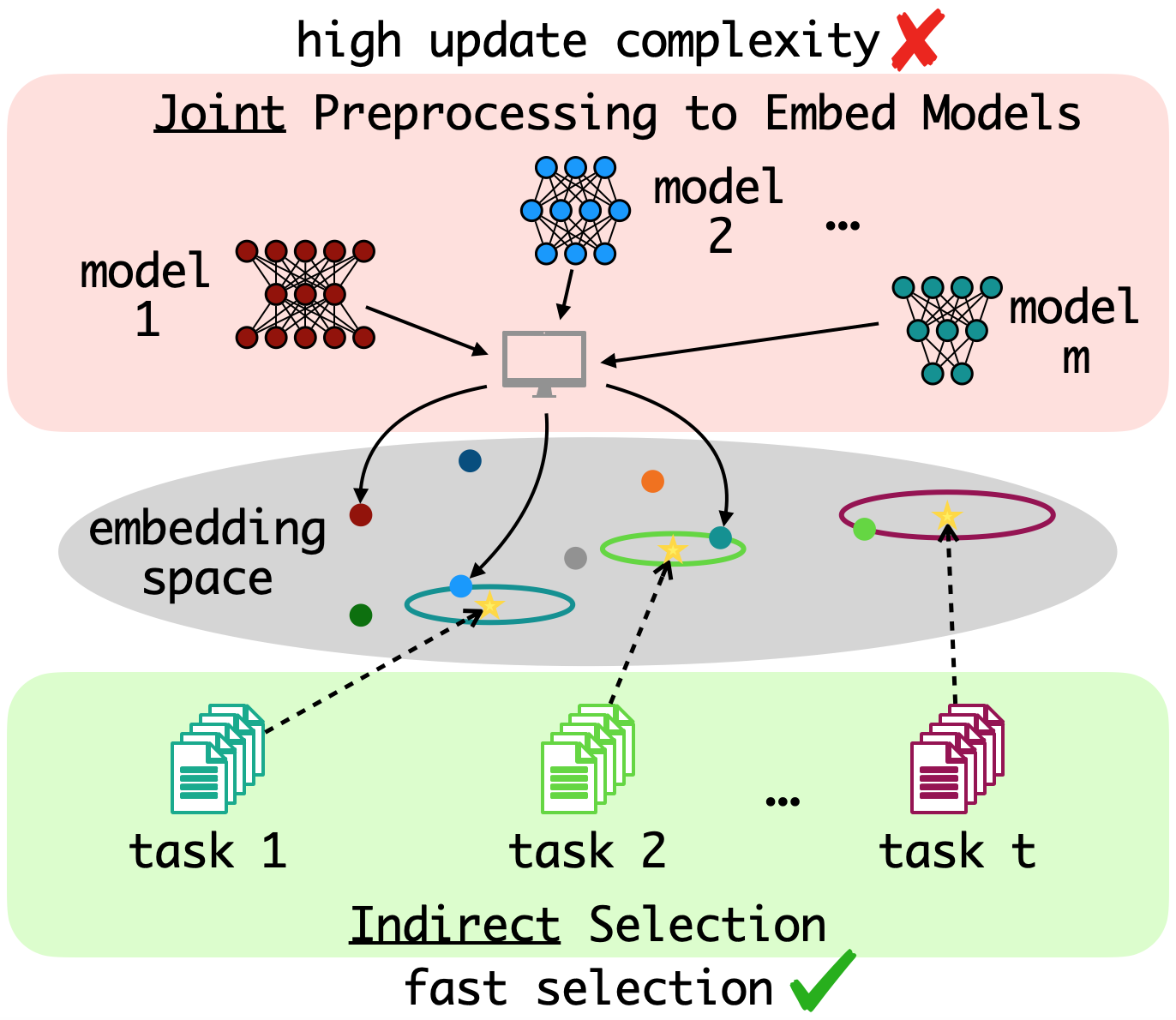}
\label{fig:QueryEfficient_Model_Selection}
}
\hfill
\subfigure[\textbf{Isolated Model Embedding}]{
\includegraphics[width=0.31\linewidth]{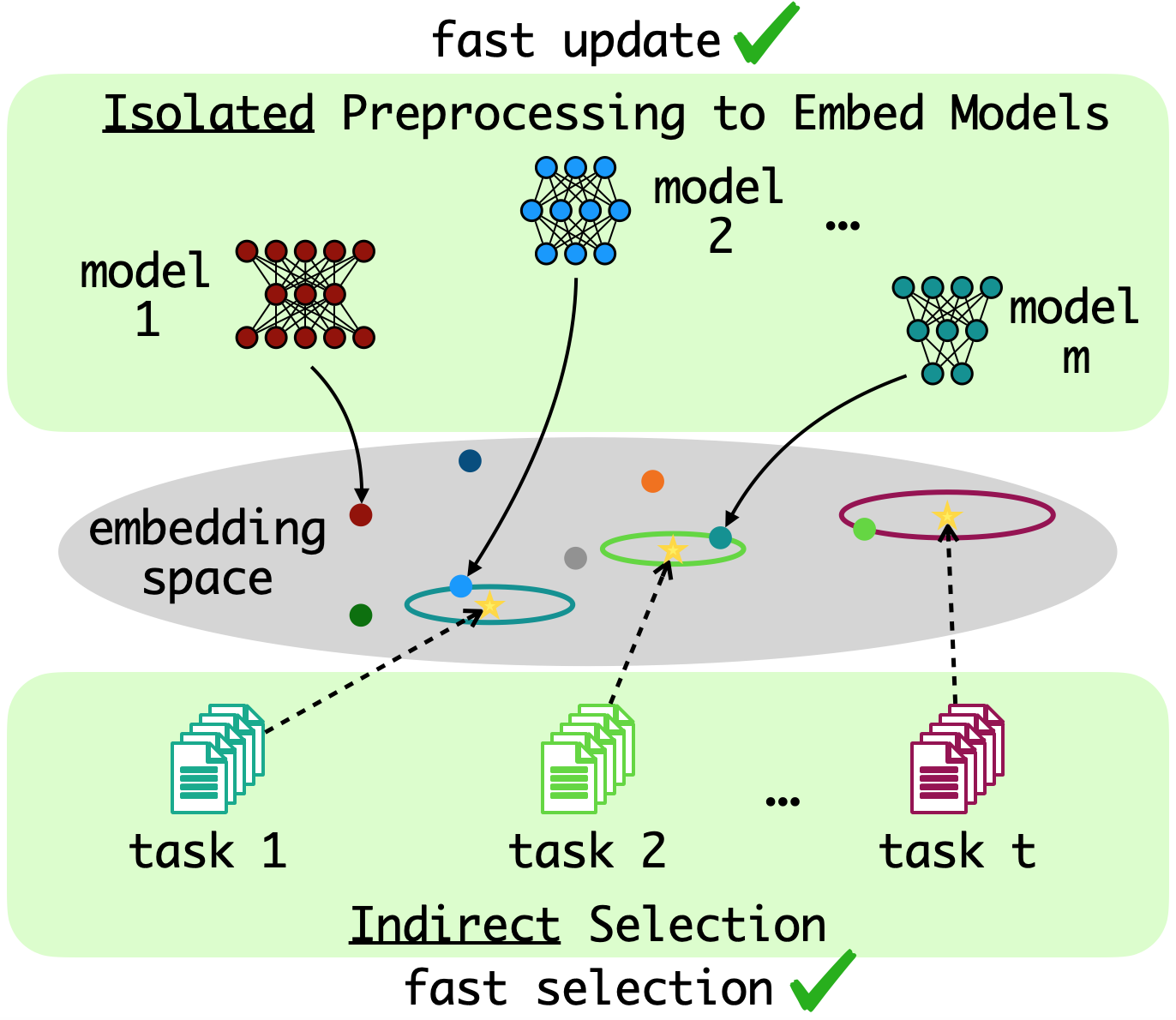}
\label{fig:IndepPrep_QueryEfficient_Model_Selection}
}
\caption{Illustrations for different families of model selection schemes. \textbf{Isolated model embedding (ours)} is a family that supports asymptotically fast update and selection \textbf{at the same time}.}
\label{fig:Formulation}
\end{center}
\vskip -0.1in
\end{figure}

New models are being created and becoming available at a rate beyond previous imaginations. Hugging Face Hub, a web platform for hosting machine learning models, datasets, and demo applications, included more than 300k pre-trained models in August 2023, when its owner, the company named Hugging Face, obtained a \$4.5 billion valuation while raising funding of \$235 million backed by 
multiple investors.\footnote{https://www.nasdaq.com/articles/ai-startup-hugging-face-valued-at-\$4.5-bln-in-latest-round-of-funding}
Such vast amounts of models will become more valuable if we can identify the ones suitable for tasks of interest, motivating the study of scalable model selection. 

Arguably, an ideal model selection scheme should minimally be able to support two operations efficiently over a large number of candidate models, which are update and selection. \textbf{Update} is an operation to either add a new candidate model or to remove an existing candidate model, which is necessary for keeping a dynamic model pool that is up-to-date. \textbf{Selection} is an operation to locate proper models for a given task, which is the core functionality of model selection. Unfortunately, existing solutions require high computational complexity for at least one of these two operations.

The most naive solution for model selection is to directly examine every candidate model on the given task for each selection operation, i.e. a brute-force solution as illustrated in Figure \ref{fig:PrepFree_Model_Selection}. Consequently, the computational complexity, measured by the number of model operations (e.g. forward/backward passes), is linear to the number of candidate models for \textbf{each selection operation}. Such complexity is prohibitively high when dealing with a large number of models and with multiple downstream tasks (i.e. multiple selection operations), even when using transferability metrics \citep{H_score, leep, LogMe, GBC, TransRate, HowStableAreMetrics, SelectBenchmark} as surrogates to reduce the computational cost of model training/tuning.

Model embedding offers a direction to improve selection complexity.
Rising from the study of task similarities \citep{zamir2018taskonomy, Task2Vec, wasserstein_task_embed, EffTuned_task_embed}, Model2Vec \citep{Task2Vec} computes an embedding vector for each candidate model, which can be used repeatedly across multiple selection operations. For each selection operation, a task embedding will first be computed using the corresponding downstream data, which takes a bounded amount of model operations that is independent from the number of candidate models, and then be used to compare with pre-computed model embeddings to locate the promising candidate models, which takes a single sweep over the model embeddings. 
As a result, with respect to the number of candidate models $m$, the number of model operations per selection is reduced to $O(1)$ with the cost of a single additional sweep over $m$ vectors.
An illustration is provided in Figure \ref{fig:QueryEfficient_Model_Selection}.

However, Model2Vec computes the model embeddings through a joint optimization that involves all candidate models, meaning that the Model2Vec embedding of each model depends on not only the model itself but also other ones in the candidate pool. This implies high complexity for each update operation that is linear to the number of candidate models (assuming no major algorithmic change) as existing model embeddings are subject to change with the addition of any new candidate model.

\textbf{In this work, we target fundamentally (more) scalable model selection that supports asymptotically fast update and asymptotically fast selection at the same time}. Specifically:

\textbf{(1)} We define \textbf{isolated model embedding}, a family of model selection schemes with asymptotically fast update and selection. Intuitively, isolated model embedding refers to a subset of the model embedding family where computing the embedding of each individual model is isolated from the others, as illustrated in Figure \ref{fig:IndepPrep_QueryEfficient_Model_Selection}.
In another word, the embedding of each model will be independent from other models in the candidate pool, which implies that each update operation involves only a single candidate model as all the previously computed model embeddings stay unchanged. 
Thus for isolated model embedding, with respect to the number of candidate models $m$, the update complexity is O(1) and the selection consists of a single sweep over $m$ vectors in addition to O(1) model operations. Notably, isolated model embedding also implies several other desirable properties.

\textbf{(2)} We present \textbf{Standardized Embedder}, an empirical realization of isolated model embedding. 
The key intuition of \name is standardization, i.e. using one public model as the baseline to embed all different candidate models, thus ensuring the independently learned embedding vectors conform to the same standard and are therefore comparable.
We use it to select representations from a pool of 100 pre-trained vision models for classification tasks and assess the performance gaps between the selected models and the best candidates with a linear probing protocol. Experiments suggest our realization is effective in selecting competitive models and highlight isolated model embedding as a promising direction towards fundamentally (more) scalable model selection. 
\section{Related Work}
\label{sec:related_work}

\textbf{Transferability metrics for model selection.} Intuitively, transferability metrics are scores that correlate with the performance of models/features after being transferred to a new task and can be computed without training: H-score \citep{H_score} is defined by incorporating the estimated inter-class variance and the redundancy of features; LEEP \citep{leep} estimates the distribution of target task label conditioned on the label of pre-trained tasks to construct a downstream classifier without training and use its performance as the metric; LogME \citep{LogMe} estimates the maximum value of the marginalized likelihood of the label given pre-trained features and uses its logarithm as the score; GBC \citep{GBC} uses class-wise Gaussians to approximate downstream samples in the pre-trained feature space so that class overlaps can be computed with Bhattacharyya coefficients to serve as a score; TransRate \citep{TransRate} estimates mutual information between features and labels by resorting to coding rate. Separate evaluations of transferability metrics are conducted by \citet{HowStableAreMetrics} and \citet{SelectBenchmark}.

\textbf{Model embedding for model selection.} Model2Vec is proposed jointly with Task2Vec by \citet{Task2Vec}. The goal of Task2Vec is to embed different tasks into a shared vector space and it does so by estimating the Fisher information matrix with respect to a probe network. To embed models, they first initialize the embedding of each model as the sum of the Task2Vec embedding $F$ of its pre-trained task (which is set to 0 if the task is unknown) and a learnable perturbation $b$. Then they learn the perturbations of all models jointly by predicting the best model given the distances of model embeddings to the task embeddings, which requires access to multiple downstream tasks in advance.
\section{A Family of Model Selection with Asymptotically Fast Update and Selection}
\label{sec:definition}

\subsection{Formal Definition}
\label{sec:formal_def}
\textbf{Isolated model embedding}, as we defined, is a family of model selection schemes consisting of two parts, as illustrated in Figure \ref{fig:IndepPrep_QueryEfficient_Model_Selection}: (1) \textbf{preprocessing}, where candidate models are converted into embedding vectors for later use in selecting models; (2) \textbf{selection}, where models are selected given downstream data (i.e. data corresponding to the downstream task) and the model embeddings generated during preprocessing.

\textbf{Preprocessing.} The preprocessing part is defined by a model embedding function of the form $\mathcal{P}(\text{model}) \to \mathcal{V}$,  where `$\text{model}$' denotes a single (candidate) model and $\mathcal{V}$ denotes an embedding space. Intuitively, $\mathcal{P}$ maps a single model into its corresponding embedding vectors.
Let $f_1, f_2, \dots, f_m$ be all current candidate models where $m$ is the total number of candidate models. For preprocessing, embedding vectors of all candidate models are generated as $\{v_i  = \mathcal{P}(f_i)\}_{i=1}^m$, where $v_i$ is the embedding vector corresponding to the candidate model $f_i$. 
Notably, this definition naturally enforces the model embedding process to be isolated: The model embedding function $\mathcal{P}$ takes a single candidate model as its input and is applied independently to each candidate model, which means the computing of each individual embedding is isolated from other models in the candidate pool. \emph{Update operations} of isolated model embedding are defined within preprocessing, where adding a new model is essentially computing its corresponding embeddings with the model embedding function $\mathcal{P}$ and removing a candidate model is removing its corresponding model embedding.

\textbf{Selection.} The selection part is defined by a task embedding function of the form $\mathcal{Q}(\text{data}) \to \mathcal{V}$ and a selection metric of the form $\delta: \mathcal{V} \times \mathcal{V} \to \mathbb{R}$, where `$\text{data}$' denotes the downstream data and $\mathcal{V}$ denotes again the embedding space. These define the \textbf{selection operations} of isolated model embedding. For each selection operation, a task embedding vector $v_{\text{task}}$ will first be computed by applying the task embedding function $\mathcal{Q}$ to the downstream data and then be compared with pre-computed model embeddings to locate promising candidate models by finding model embeddings $v_i$ that maximize the selection metric  $\delta(v_\text{task}, v_i)$.

\subsection{Asymptotically Fast Update and Selection}
\label{sec:complexity}
\begin{table*}[tbp!]
\caption{A summary of the computational complexity per operation for different families of model selection schemes with respect to the (current) number of candidate models $m$. *Selection complexity is measured by the number of model operations, e.g. forward/backward passes of models.}
\label{tab:property}
\renewcommand*{\arraystretch}{1.08}
\begin{center}
\begin{tabular}{c||cc}
Family  & Update Complexity 
& Selection Complexity* \\ 
\hline \hline
Brute-force &   O(1)  &    O(m)       \\
Model embedding  &   O(m)  &   O(1)        \\ \hline
\multirow{1}{*}{\shortstack[c]{ \textbf{Isolated model embedding}}}                  &   \textbf{O(1)}   &
\multirow{1}{*}{\textbf{O(1)}}            
\end{tabular}
\end{center}
\vskip -0.1in
\end{table*}

In this section, we show how isolated model embedding supports asymptotically fast update and selection by analyzing computational complexity. A summary of update complexity and selection complexity for different families of model selection schemes is included in Table \ref{tab:property}.

\textbf{Update complexity:} An update operation either adds a new candidate model or removes an existing one. With isolated model embedding, adding a new model is essentially computing its corresponding embeddings with the model embedding function $\mathcal{P}$. Since the model embedding function $\mathcal{P}$ takes a single model as its input and has no dependency on the number of the candidate models $m$, the computational complexity of $\mathcal{P}$ (and therefore the complexity of adding a new model) must be O(1) with respect to $m$.
Removing a candidate model is simply removing its corresponding embedding vectors from a maintained list containing embeddings of all the (current) candidate models, which can also be O(1). Thus with respect to the number of candidate models $m$, the complexity per update operation is O(1) for isolated model embedding.

\textbf{Selection complexity:} With isolated model embedding, a selection operation consists of two steps. The first step is to use the task embedding function $\mathcal{Q}$ to compute a task embedding vector $v_{\text{task}}$. The task embedding function $\mathcal{Q}$ takes the downstream task data as the input and and has no dependency on the number of the candidate models $m$, thus the computational complexity of $\mathcal{Q}$ is O(1) with respect to $m$, which also means that there can be at most O(1) model operations.
The second step is to compare the task embedding $v_{\text{task}}$ with the model embeddings $\{v_i\}_{i=1}^m$ by computing the selection metric $\delta(v_\text{task}, v_i)$, which takes a single sweep over $m$ model embedding vectors. As a result, with respect to the number of candidate models $m$, each selection operation consists of O(1) model operations in addition to a single sweep over $m$ embedding vectors. 
Notably, while the selection complexity is technically still O(m), reducing the number of model operations from O(m) to O(1) greatly improves scalability, as model operations (e.g. forward/backward passes) are typically orders of magnitude slower than vector operations (e.g. inner products) in practice.

\subsection{Other Desirable Properties}
\label{sec:implication}

\textbf{Isolated model embedding is naturally decentralizable.} 
Model owners can make their models candidates of future model selections entirely on their own, requiring no collaboration with other model owners and requiring no centralized coordination.
Given the model embedding function $\mathcal{P}$, model owners can independently embed their models and publish/broadcast the resulting embeddings by themselves. After that, any party can use the task embedding function $\mathcal{Q}$ and the selection metric $\delta$ to select models for its own tasks, using all the model embeddings that it has access to. This is a quite simple decentralized protocol from isolated model embedding.

\textbf{Embedding vectors are flexible and portable information carriers.} 
Model embeddings are simply (real) vectors, which can be stored and processed in diverse formats through various packages, making it a bridge connecting different implementation frameworks, different owners and different platforms. For example, some model owners may use PyTorch for their models and for implementing the model embedding function $\mathcal{P}$ while some model owners may use TensorFlow, but this creates little difficulty for a selection operation even if it is implemented in neither frameworks since the selection operation only needs to recognize the computed embedding vectors.
In addition, typical dimensions of model embeddings are fairly small and therefore the sizes are quite portable: For instance, in later empirical evaluations, we incorporate settings with 512-dimensional and 768-dimensional model embeddings, while as an informal comparison, each of Figure \ref{fig:PrepFree_Model_Selection}, \ref{fig:QueryEfficient_Model_Selection} and \ref{fig:IndepPrep_QueryEfficient_Model_Selection} contains $1372\times 1190 \approx 1.6\times 10^6$ RGB pixels, which is about $4.8\times 10^6$ dimensions with 8 bits each. Thus a single image of such has already enough bits to encode the embedding vectors as 32-bit floats for more than $1000$ models.

\textbf{Candidate models can be kept private throughout model selection.} Model owners can embed their models using the model embedding function $\mathcal{P}$ by themselves, 
and the only model-dependent information required by selection operations is the model embeddings. Thus, for the entire model selection process, model owners do not need to release their models to any other parties, including but not limited to other model owners and downstream users who want to select models for their downstream tasks.
As selection completes, downstream users can reach out to the owners of selected candidates (instead of all model owners) for requesting access.

\textbf{Selection operations can be made invisible to model owners.} Given the model embeddings, selection operations can be performed fully locally by downstream users. Consequently, downstream data can be kept private and model owners will have no knowledge regarding the selection or whether there is a selection, unless the downstream users choose to notify them. This can be a highly valuable privacy guarantee for downstream users.

\section{\name: A Realization of Isolated Model Embedding}
\label{sec:method}

In this section, we present \name, an empirical realization of isolated model embedding. 
As defined in Section \ref{sec:definition} and illustrated in Figure \ref{fig:IndepPrep_QueryEfficient_Model_Selection}, isolated model embedding contains two parts: \textbf{preprocessing}, where model embedding vectors are learned independently for different candidates, and \textbf{selection}, where a task embedding vector is learned from the downstream data and is used to search among model embeddings to guide model selection. We will introduce some concepts as tools in Section \ref{sec:tool} before we present these two parts respectively in Section \ref{sec:preparation} and \ref{sec:query}. 

\subsection{Tool: (Approximate) Functionality Equivalence}
\label{sec:tool}

Firstly, we introduce notations. Let $\mathcal{X}$ be the input space. A feature $f: \mathcal{X} \to \mathbb{R}$ is defined as a function mapping any sample from the input space $\mathcal{X}$ to a real number. A set of features $F$ is therefore a set of functions, which can also be considered as a function $F:\mathcal{X}\to \mathbb{R}^n$ mapping any sample to a vector of $n$ dimensions, where $n$ can be either finite or countably infinite, depending on whether or not the set contains a finite number of features. 

\begin{definition}[Functionality Equivalence]
    For two sets $F: \mathcal{X} \to \mathbb{R}^n$ and $\hat{F}: \mathcal{X} \to \mathbb{R}^m$ of features, they are considered $\delta$-equivalent in functionality over a distribution $D$ over $\mathcal{X}$, if and only if there exist two affine transformations $w,b \in \mathbb{R}^{n\times m} \times \mathbb{R}^m$ and $\hat{w}, \hat{b}\in \mathbb{R}^{m\times n} \times \mathbb{R}^n$ such that
    \begin{align*}
            &\E_{x\sim D}\left[ S_\text{cos}\left(w^\top F(x) + b, \hat{F}(x)\right) \right] \geq 1 - \delta
            \text{~~and~~}
            \E_{x\sim D}\left[ S_\text{cos}\left(F(x), \hat{w}^\top \hat{F}(x) + \hat{b}\right) \right] \geq 1 - \delta,
    \end{align*}
    where $S_\text{cos}\left(u, v\right)$ denotes cosine similarity between two vectors $u$ and $v$.
    \label{def:equivalence}
\end{definition}

Functionality equivalence characterizes cases where two sets of features are considered the same regarding their usability in unknown applications. 
Intuitively, since most (if not all) modern architectures of neural networks have at least one affine transformation following the feature layers, two sets of features should be considered equivalent even if they differ by an affine transformation. Similar arguments are introduced by \cite{wang2018towards} to understand deep representations, where they consider two representations to be equivalent when the subspaces spanned by their activation vectors are identical. While in principle other similarity metrics can be utilized as well, we use cosine similarity in this work since it is naturally invariant to feature scalings.

\begin{figure*}[t!]
\begin{center}
\subfigure[Preprocessing: isolated model embedding]{
\includegraphics[width=0.625\linewidth]{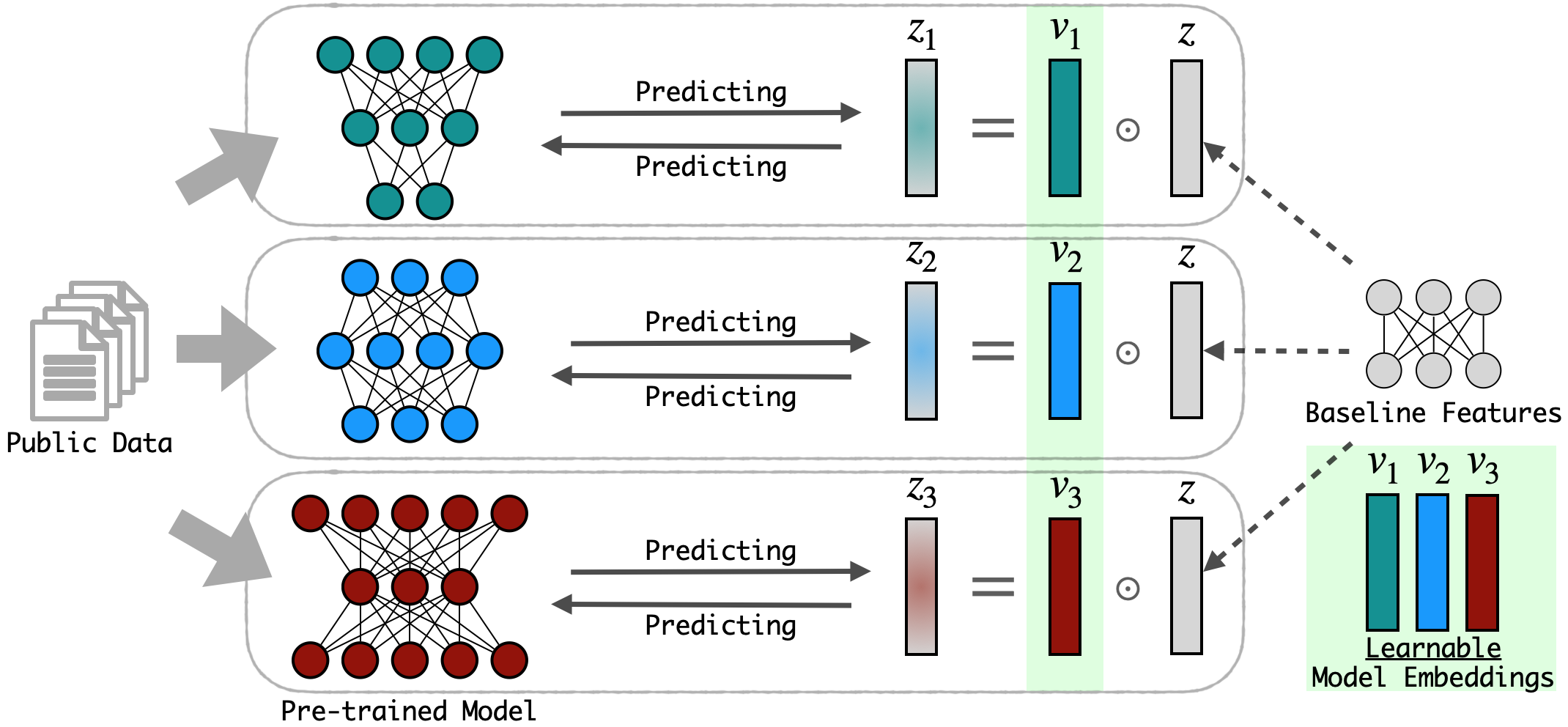}
\label{fig:ViewingModelAsVectors}
}
\hfill
\subfigure[Selection: task embedding \& \newline selection metric]{
\includegraphics[width=0.325\linewidth]{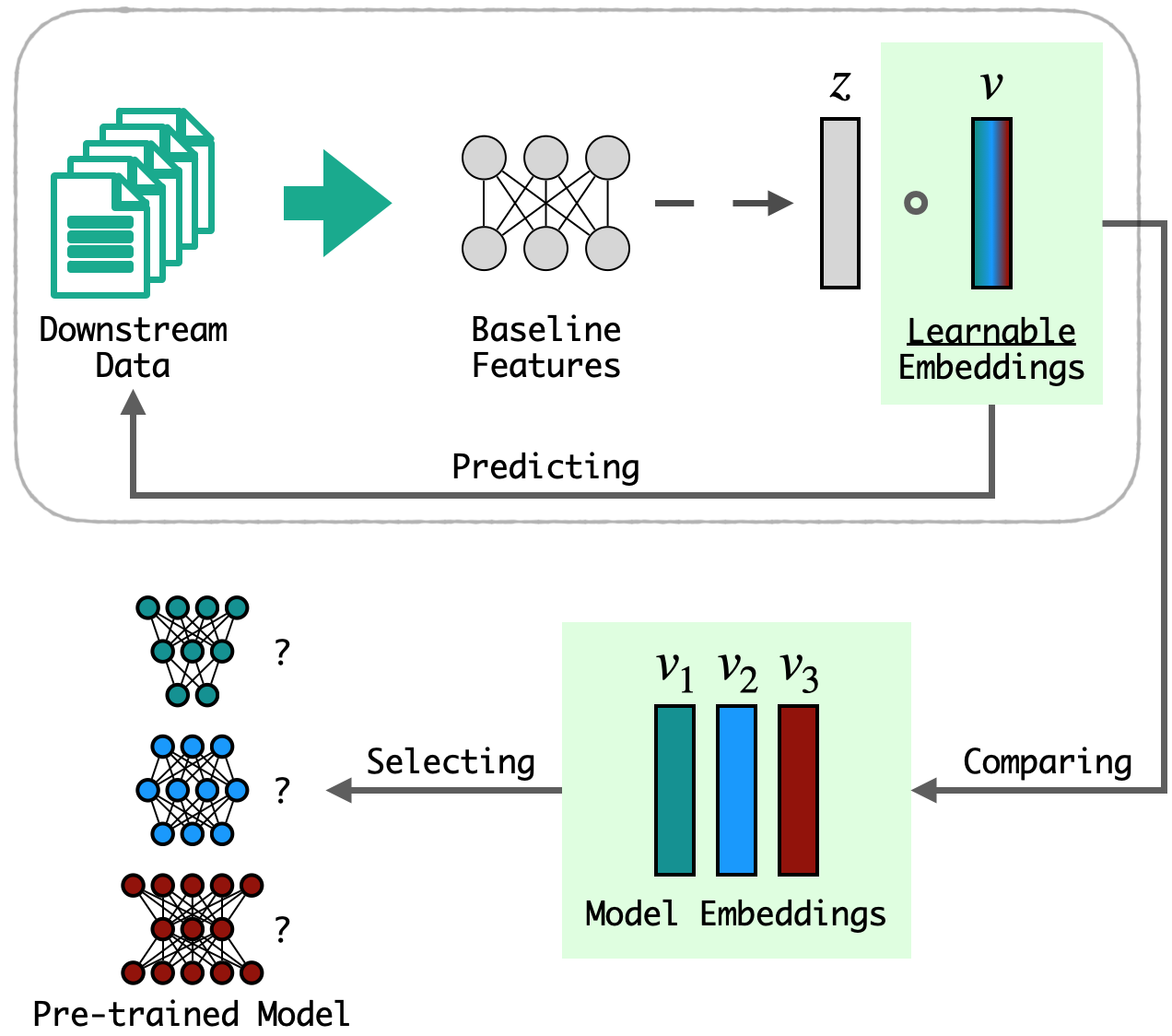}
\label{fig:SelectingModelAsVectors}
}
\caption{An illustration of \name. \textbf{(a)~Preprocessing:} Using features of a public model as the baseline, a vector embedding is learned \textbf{independently} for each pre-trained model. Intuitively, the embeddings of models denote their approximately equivalent feature subsets in the baseline features. \textbf{(b)~Selection:} Task embeddings are defined by subsets of the baseline features that are important to corresponding downstream tasks, which are identified through enforcing sparsity regularization. Models are selected by comparing the task embedding with model embeddings of all candidates, using (the cardinality of) standard fuzzy set intersection as the selection metric.}
\end{center}
\vskip -0.1in
\end{figure*}

\subsection{Preprocessing: Isolated Model Embedding by Identifying Equivalent Feature Subsets}
\label{sec:ViewingModelAsVectors}
\label{sec:preparation}

With functionality equivalence from Definition \ref{def:equivalence}, we can characterize the representation powers of any set of features by associating them with the equivalent subsets from a pre-defined, baseline feature set $B: \mathcal{X} \to \mathbb{R}^N$ (Empirically we will use a public model as this baseline feature set, which will be elaborated in Section \ref{sec:eval}). Since any subset of the baseline feature set $B: \mathcal{X} \to \mathbb{R}^N$ can be directly associated with a binary vector from $\{0, 1\}^N$ (i.e. each $1$ indicating the presence of a feature and each $0$ indicating an absence; See Appendix \ref{appendix:toyexample} for examples), we simply use such vectors as the embeddings of models. For the actual implementation, we relax this binary embedding space to a continuous one (i.e. $[0, 1]^N$). The formal definition is as follows.

\begin{definition}[Vector Embedding through Equivalent Feature Subsets]
Given a baseline feature set $B: \mathcal{X} \to \mathbb{R}^N$, a vector $v \in \{0, 1\}^N$ is a $\delta$-embedding vector of a feature set $F: \mathcal{X} \to \mathbb{R}^n$ over a distribution $D$ if and only if $F$ and $\{B_i| v_i = 1\}$ are $\delta$-equivalent in functionality over $D$.
\label{def:embed_subset}
\end{definition}

Given a set of features as the baseline, the embedding vectors corresponding to a set of features are defined through Definition \ref{def:embed_subset}.
Consequently, we can now conceptually map each model, represented as a set of features, into a vector embedding space that associates with the baseline features.

In practice, to compute the embedding vectors given baseline features, we relax the binary embedding space to a continuous one and reformulate it as follows:
\begin{align*}
    &\max_{v, w, b, \hat{w}, \hat{b}}~\min\left(L_\text{to baseline}, L_\text{from baseline}\right)\\
    \text{subject to:}~
    &L_\text{to baseline} = \E_{x\sim D} \left[ S_\text{cos}\left(w^\top F(x) + b, v\odot B(x)\right) \right]\\
    &L_\text{from baseline} = \E_{x\sim D} \left[ S_\text{cos}\left(F(x), \hat{w}^\top \left(v\odot B(x)\right) + \hat{b}\right) \right]
    \\
    &v\in [0,1]^n,  w \in \mathbb{R}^{n\times N}, b\in \mathbb{R}^N, \hat{w} \in \mathbb{R}^{N\times n}, \hat{b}\in \mathbb{R}^n
\end{align*}
where $F: \mathcal{X} \to \mathbb{R}^n$ is the feature set that we want to vectorize, $B: \mathcal{X} \to \mathbb{R}^N$ is the set of baseline features, $D$ is the underlying data distribution, $v$ is the (relaxed) embedding vector, $w,b,\hat{w},\hat{b}$ are parameters of affine transforms and $\odot$ denotes Hadamard product (i.e. element-wise multiplication). An illustration of the model embedding process is included in Figure \ref{fig:ViewingModelAsVectors}.

Empirically, the constraint $v\in[0,1]^n$ is implemented via reparameterization through the sigmoid function, i.e. $v_i = 1/(1 + e^{-v'_i / \tau})$, where $\tau$ is a constant known as temperature and we use a fixed temperature of $\tau=0.01$ in all experiments. Intuitively, the optimization wants to find a subset of the baseline features (indicated by the mask $v$) that is $\delta$-equivalent to $F$ for smaller $\delta$.

Both Definition \ref{def:embed_subset} and the relaxation are straightforward, but it is worth noting that the embedding depends on not only the model (i.e. the set of features) to be embedded, but also the set of baseline features, and the embedding vectors may not be unique by definition depending on the choice of baseline features. 
Conceptually, what we do here is to compare the embedding distributions of different models by drawing a single embedding vector from each distribution.

\subsection{Selection: Task Embedding through Feature Sifting}
\label{sec:query}

In this section, we showcase how to derive a task embedding vector from downstream data and the selection metric for comparing the task embedding with model embeddings. 
An illustration of the process is included in Figure \ref{fig:SelectingModelAsVectors}: Intuitively, we derive the task embedding vector by identifying subsets of the baseline features $B: \mathcal{X}\to \mathbb{R}^N$ that are important to the task of interest, which can then be directly associated with binary vectors from $\{0,1\}^N$, similar to how we previously embed models as vectors, and we use a measure of (fuzzy) set similarity as the selection metric.

Formally, for a downstream task, let $\mathcal{X}$ be the input space, $\mathcal{Y}$ be the label space, we use $\hat{D}$ to denote the downstream data distribution, which is a distribution over $\mathcal{X}\times \mathcal{Y}$. Using $L$ to denote the corresponding task loss, identifying important features can be formulated as follows:
\begin{align*}
    \min_{v, w, b}~ & \E_{x,y\sim \hat{D}} \left[ L\left(w^\top (v \odot B(x)) + b, y\right) \right] + \gamma  ||v||_1 &
    \text{subject to:}~& ||w^\top||_1 = 1 \text{ and } v\in [0,1]^n
\end{align*}
where $v\in [0,1]^n$ is the embedding vector of the task to be learned, $w,b\in \mathbb{R}^{n\times |\mathcal{Y}|} \times \mathbb{R}^{|\mathcal{Y}|}$ jointly denotes a prediction head associated with the task of interest, $||v||_1$ denotes the $\ell_1$ norm of the embedding vector (which functions as sparsity regularization), $||w^\top||_1$ denotes the matrix norm of $w^\top$ induced by $\ell_1$ norm of vectors (i.e. $||w^\top||_1 = \sup_{x\neq 0} ||w^\top x||_1 / ||x||_1 = \max_i \sum_j |w_{ij}|$) and $\gamma$ is sparsity level, a scalar hyper-parameter controlling the strength of the sparsity regularization $\gamma ||v||_1$. A rule of thumb for choosing $\gamma$ is suggested in Section \ref{sec:eval_choice_L1reg}.

\textbf{Selection metric.} Given the task embedding, we compare it with embeddings of candidate models to identify the most similar ones to the task embedding with respect to a similarity metric----the corresponding models are the ones to be selected.
Notably, all our embedding vectors, including model embeddings and task embeddings, are relaxations of binary vectors denoting subsets of the baseline features. This is well related to fuzzy set theory \citep{zadeh1965fuzzy, klir1995fuzzy} where each element can have a degree of membership between 0 and 1 to indicate whether it is not/fully/partially included in a set. Interpreting both model embeddings and task embeddings as fuzzy sets, we incorporate standard fuzzy set intersection to measure the similarity between model embeddings and task embeddings. Formally, let $u, v\in [0,1]^n$ be two embedding vectors (interpreted as fuzzy sets), the cardinality of their standard intersection is simply
 $   I_\text{standard}(u, v) = \sum_{i=1}^n \min(u_i, v_i).$

Intuitively, the task embedding denotes the set of baseline features useful for the task of interest and each model embedding denotes the set of baseline features possessed by the corresponding candidate model. Thus the cardinality of their intersection measures the quantity of the useful features owned by candidate models and therefore provides guidance for downstream performance.

\section{Evaluation}
\label{sec:eval}

\begin{table*}[t!]
\renewcommand{\arraystretch}{1.03} 
\caption{Empirical evaluations of \name with \textbf{100 pre-trained models} as the candidates (See Appendix \ref{ap:candidates} for the full list). \name successfully locates models comparable to the best candidates for corresponding downstream tasks.}
\label{tab:selection_results}
\begin{center}
\resizebox{0.85\linewidth}{!}{
\begin{tabular}{|c|c||c|c|ccc|}
\hline
\multirow{2}{*}{Downstream task} &
  \multirow{2}{*}{\begin{tabular}[c]{@{}c@{}}Best candidate\\ \textbf{(Ground truth)}\end{tabular}} &
  \multirow{2}{*}{\begin{tabular}[c]{@{}c@{}}Model used as \\baseline features\end{tabular}} &
  \multirow{2}{*}{\begin{tabular}[c]{@{}c@{}}Training steps\\ per candidate\end{tabular}} &
  \multicolumn{3}{c|}{\begin{tabular}[c]{@{}c@{}}Downstream accuracy of selected models\\ \textbf{(+gap from the best candidate)}\end{tabular}} \\ \cline{5-7} 
 &
   &
   &
   &
  \multicolumn{1}{c|}{best of top 1} &
  \multicolumn{1}{c|}{best of top 3} &
  best of top 5 \\ \hline \hline
\multirow{7}{*}{CIFAR-10} &
  \multirow{7}{*}{95.15\%} &
  \multirow{3}{*}{ResNet-18} &
  4k &
  \multicolumn{1}{c|}{\begin{tabular}[c]{@{}c@{}}91.81\%\\ \textbf{(3.34\%)}\end{tabular}} &
  \multicolumn{1}{c|}{\begin{tabular}[c]{@{}c@{}}94.57\%\\ \textbf{(0.58\%)}\end{tabular}} &
  \begin{tabular}[c]{@{}c@{}}94.57\%\\ \textbf{(0.58\%)}\end{tabular} \\ \cline{4-7} 
 &
   &
   &
  10k &
  \multicolumn{1}{c|}{\begin{tabular}[c]{@{}c@{}}94.36\%\\ \textbf{(0.79\%)}\end{tabular}} &
  \multicolumn{1}{c|}{\begin{tabular}[c]{@{}c@{}}95.12\%\\ \textbf{(0.03\%)}\end{tabular}} &
  \begin{tabular}[c]{@{}c@{}}95.12\%\\ \textbf{(0.03\%)}\end{tabular} \\ \cline{3-7} 
 &
   &
  \multirow{3}{*}{Swin-T (tiny)} &
  4k &
  \multicolumn{1}{c|}{\begin{tabular}[c]{@{}c@{}}94.57\%\\ \textbf{(0.58\%)}\end{tabular}} &
  \multicolumn{1}{c|}{\begin{tabular}[c]{@{}c@{}}94.57\%\\ \textbf{(0.58\%)}\end{tabular}} &
  \begin{tabular}[c]{@{}c@{}}95.12\%\\ \textbf{(0.03\%)}\end{tabular} \\ \cline{4-7} 
 &
   &
   &
  10k &
  \multicolumn{1}{c|}{\begin{tabular}[c]{@{}c@{}}95.12\%\\ \textbf{(0.03\%)}\end{tabular}} &
  \multicolumn{1}{c|}{\begin{tabular}[c]{@{}c@{}}95.12\%\\ \textbf{(0.03\%)}\end{tabular}} &
  \begin{tabular}[c]{@{}c@{}}95.12\%\\ \textbf{(0.03\%)}\end{tabular} \\ \hline \hline
\multirow{7}{*}{CIFAR-100} &
  \multirow{7}{*}{82.58\%} &
  \multirow{3}{*}{ResNet-18} &
  4k &
  \multicolumn{1}{c|}{\begin{tabular}[c]{@{}c@{}}81.11\%\\ \textbf{(1.47\%)}\end{tabular}} &
  \multicolumn{1}{c|}{\begin{tabular}[c]{@{}c@{}}81.11\%\\ \textbf{(1.47\%)}\end{tabular}} &
  \begin{tabular}[c]{@{}c@{}}81.11\%\\ \textbf{(1.47\%)}\end{tabular} \\ \cline{4-7} 
 &
   &
   &
  10k &
  \multicolumn{1}{c|}{\begin{tabular}[c]{@{}c@{}}80.13\%\\ \textbf{(2.45\%)}\end{tabular}} &
  \multicolumn{1}{c|}{\begin{tabular}[c]{@{}c@{}}80.13\%\\ \textbf{(2.45\%)}\end{tabular}} &
  \begin{tabular}[c]{@{}c@{}}81.57\%\\ \textbf{(1.01\%)}\end{tabular} \\ \cline{3-7} 
 &
   &
  \multirow{3}{*}{Swin-T (tiny)} &
  4k &
  \multicolumn{1}{c|}{\begin{tabular}[c]{@{}c@{}}81.11\%\\ \textbf{(1.47\%)}\end{tabular}} &
  \multicolumn{1}{c|}{\begin{tabular}[c]{@{}c@{}}81.48\%\\ \textbf{(1.10\%)}\end{tabular}} &
  \begin{tabular}[c]{@{}c@{}}81.48\%\\ \textbf{(1.10\%)}\end{tabular} \\ \cline{4-7} 
 &
   &
   &
  10k &
  \multicolumn{1}{c|}{\begin{tabular}[c]{@{}c@{}}81.48\%\\ \textbf{(1.10\%)}\end{tabular}} &
  \multicolumn{1}{c|}{\begin{tabular}[c]{@{}c@{}}81.48\%\\ \textbf{(1.10\%)}\end{tabular}} &
  \begin{tabular}[c]{@{}c@{}}81.48\%\\ \textbf{(1.10\%)}\end{tabular} \\ \hline \hline
\multirow{7}{*}{STL-10} &
  \multirow{7}{*}{99.24\%} &
  \multirow{3}{*}{ResNet-18} &
  4k &
  \multicolumn{1}{c|}{\begin{tabular}[c]{@{}c@{}}98.76\%\\ \textbf{(0.47\%)}\end{tabular}} &
  \multicolumn{1}{c|}{\begin{tabular}[c]{@{}c@{}}98.76\%\\ \textbf{(0.47\%)}\end{tabular}} &
  \begin{tabular}[c]{@{}c@{}}98.76\%\\ \textbf{(0.47\%)}\end{tabular} \\ \cline{4-7} 
 &
   &
   &
  10k &
  \multicolumn{1}{c|}{\begin{tabular}[c]{@{}c@{}}97.51\%\\ \textbf{(1.72\%)}\end{tabular}} &
  \multicolumn{1}{c|}{\begin{tabular}[c]{@{}c@{}}98.60\%\\ \textbf{(0.64\%)}\end{tabular}} &
  \begin{tabular}[c]{@{}c@{}}98.76\%\\ \textbf{(0.47\%)}\end{tabular} \\ \cline{3-7} 
 &
   &
  \multirow{3}{*}{Swin-T (tiny)} &
  4k &
  \multicolumn{1}{c|}{\begin{tabular}[c]{@{}c@{}}97.69\%\\ \textbf{(1.55\%)}\end{tabular}} &
  \multicolumn{1}{c|}{\begin{tabular}[c]{@{}c@{}}98.60\%\\ \textbf{(0.64\%)}\end{tabular}} &
  \begin{tabular}[c]{@{}c@{}}98.60\%\\ \textbf{(0.64\%)}\end{tabular} \\ \cline{4-7} 
 &
   &
   &
  10k &
  \multicolumn{1}{c|}{\begin{tabular}[c]{@{}c@{}}98.60\%\\ \textbf{(0.64\%)}\end{tabular}} &
  \multicolumn{1}{c|}{\begin{tabular}[c]{@{}c@{}}98.60\%\\ \textbf{(0.64\%)}\end{tabular}} &
  \begin{tabular}[c]{@{}c@{}}98.60\%\\ \textbf{(0.64\%)}\end{tabular} \\ \hline
\end{tabular}
}
\end{center}
\vskip -0.1in
\end{table*}

\subsection{Evaluation Setup}
\label{sec:eval_setup}

For evaluations, we gather \textbf{100 pre-trained models} with various architectures and training recipes as candidates of model selection. See Appendix \ref{ap:candidates} for the full list of models. 

\textbf{Preprocessing.} Recalling that \name uses a public model as the baseline features and (approximate) functionality equivalence is defined over an underlying distribution D, we include two choices of baseline features, a pre-trained ResNet-18 or a pre-trained Swin Transformer (tiny) and we use the validation set of ImageNet (50000 samples in total) as the underlying distribution D (for the embedding of candidate models). We use the same hyper-parameters when embedding different candidate models: We use SGD optimizer with a batch size of 128, an initial learning rate of 0.1, a momentum of 0.9, and a weight decay of 5e-4; We divide the learning rate by a factor of 10 every 1k steps for settings with 4k training steps (=10.24 epochs) per candidate, and every 3k steps for settings with 10k training steps (=25.6 epochs) per candidate.

\textbf{Selection.} We assess the quality of representations selected for three downstream benchmarks, CIFAR-10 \citep{CIFAR}, CIFAR-100 \citep{CIFAR} and STL-10 \citep{STL10}, which are natural object classifications benchmarks with varying granularity and varying domain shifts (compared to ImageNet validation set that we use as the distribution to embed models). To learn task embeddings, we use SGD optimizer with a batch size of 128, an initial learning rate of 0.1, a momentum of 0.9, and a weight decay of 5e-4 for 60 epochs, with the learning rate getting divided by a factor of 10 every 15 epochs. Notably, the weight decay is disabled for the task embedding to be learned to avoid interfering with the sparsity regularization from Section \ref{sec:query}. For coherence, we discuss the choice of the sparsity level $\gamma$ in Section \ref{sec:eval_choice_L1reg}.

\textbf{Evaluation.} 
To estimate performance gaps between the selected models and the best candidates, 
we incorporate a linear probing protocol commonly used to assess the quality of representations \citep{chen2020simple, Hua_2021_ICCV, temporal_robustness, chen2024deconstructing}. For every candidate model and for every downstream task, a linear head is trained over its features to compute the corresponding downstream accuracy. We use SGD with a batch size of 128, an initial learning rate of 0.1, a momentum of 0.9, and a weight decay of 5e-4 for 60 epochs, with the learning rate getting divided by a factor of 10 every 15 epochs.

\subsection{The Performance of \name}
\label{sec:eval_performance}

In Table \ref{tab:selection_results}, we include the quantitative evaluations on the performance of \name. For each downstream task, we report in the left half of the table the downstream accuracy of the best candidates (i.e. the ground truth) for references and report in the right half of the table the downstream accuracy of models selected (i.e. the top-1/top-3/top-5 candidates according to the selection metric). \name successfully locates models/representations comparable to the best possible candidates with respect to different downstream tasks: When selecting only 1 model from the 100 candidates, the \textbf{worst} accuracy gap \textbf{across all evaluated settings} of \name (i.e. with different baseline features and training steps per candidate) and all downstream tasks evaluated is 3.34\%, which is reduced to 2.45\% when selecting 3 models and 1.47\% when selecting 5 models; When using Swin Transformer (tiny) as baseline features and 10k embedding steps per candidate, the worst gap is only 1.10\% even when selecting only 1 model.

In Figure \ref{fig:acc_vs_si_tmp1} and \ref{fig:acc_vs_si_tmp2}, we present both the downstream accuracy and the cardinality of standard intersection (i.e. the selection metric) for different downstream tasks when using different baseline features. More results are included in Figure \ref{fig:acc_vs_si} and \ref{fig:acc_vs_si_10k} in Appendix. We use the \textcolor{olive}{dashed line} to highlight the downstream accuracy of the public model used as baseline features. An important observation here is that \name is able to locate much more competitive models when the public model used as baseline features is only suboptimal (i.e. the public model used as baseline features performs considerably worse on the downstream tasks compared to the best possible candidate).

\subsection{On the Choice of Baseline Features}
\label{sec:eval_choice_baseline_ftrs}

\begin{figure}[t!]
\begin{center}
\subfigure[baseline features:\newline ResNet-18]{
\includegraphics[width=0.21\linewidth]{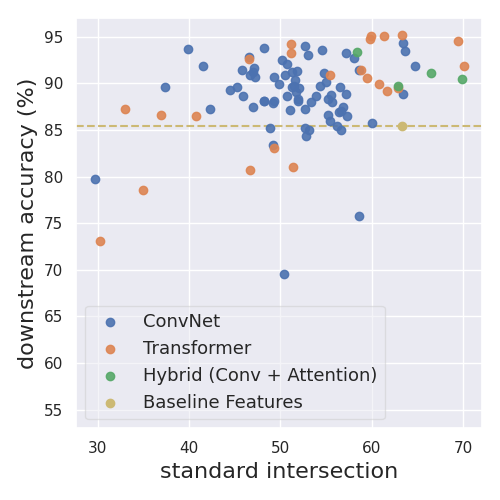}
\label{fig:acc_vs_si_tmp1}
}
\hfill
\subfigure[baseline features:\newline Swin-T (tiny)]{
\includegraphics[width=0.21\linewidth]{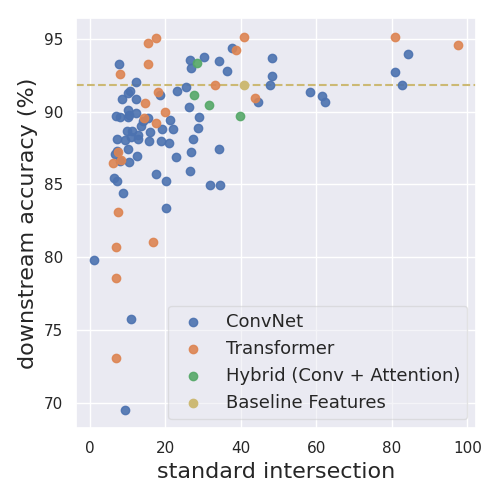}
\label{fig:acc_vs_si_tmp2}
}
\hfill
\subfigure[Choice of baseline features]{
\includegraphics[width=0.21\linewidth]{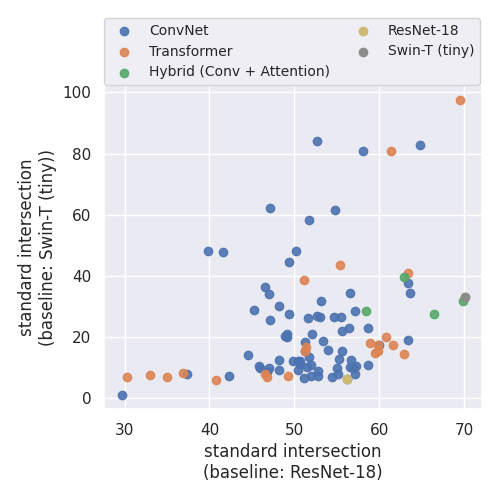}
\label{fig:si_vs_si_4k_tmp}
}
\hfill
\subfigure[Choosing sparsity level]{
\includegraphics[width=0.26\linewidth]{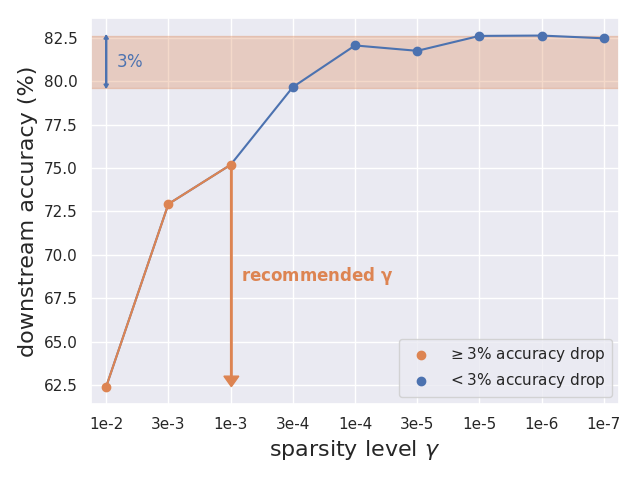}
\label{fig:choice_of_L1reg_tmp}
}
\caption{\textbf{(a, b)} Downstream accuracy (i.e. the ground truth) on CIFAR-10 v.s. the cardinality of standard intersections (i.e. the selection metric) when using 4k steps per candidate. The downstream accuracy of the baseline features are highlighted with the \textcolor{olive}{dashed line}. When a public model is only suboptimal, using it as baseline features for \name can still locate more competitive models. \textit{See Figure \ref{fig:acc_vs_si} and \ref{fig:acc_vs_si_10k} in Appendix for more results including other downstreams and more steps.} \textbf{(c)}~Comparing the cardinality of standard intersections (i.e. the selection metric) when using different baseline features (ResNet-18 and Swin-T (tiny)) with 4k steps per candidate and CIFAR-10 as the downstream task. The green/orange points in the bottom right suggest using ResNet-18 as baseline features tend to overestimate (some) models with attentions compared to using Swin Transformer (tiny). \textit{See Figure \ref{fig:si_vs_si_4k} and \ref{fig:si_vs_si_10k} in Appendix for more results including other downstreams and more steps.} \textbf{(d)}~Downstream accuracy on CIFAR-10 of the baseline features ResNet-18 corresponding to varying sparsity regularization $\gamma$. A rule of thumb for deciding the value of $\gamma$: using the smallest $\gamma$ with at least 3\% accuracy drop from the converged accuracy. \textit{See Figure \ref{fig:choice_of_L1reg} in Appendix for more results.}}

\end{center}
\vskip -0.1in
\end{figure}

Table \ref{tab:selection_results} suggests that \name performs better when using Swin Transformer (tiny) as baseline features compared to when using ResNet-18. To further understand this, we compare the selection metric of candidate models on each of the evaluated downstream tasks when using different baseline features in Figure \ref{fig:si_vs_si_4k_tmp}, with more results included in Figure \ref{fig:si_vs_si_4k} and Figure \ref{fig:si_vs_si_10k} in Appendix, where in all cases there are clusters of green/orange points in the bottom right, which corresponds to Transformers and hybrid models (i.e. models with both convolution and attention) that are overrated when using ResNet-18 as baseline features.
This is likely because the limited receptive fields of ConvNets prevents them from capturing some long-range correlations utilized by models with attention (i.e. Transformers and hybrid models),
which suggests models with attention are currently better choices of baseline features than ConvNets.

A possible limitation of \name is the robustness of the choice of baseline features to primary paradigm shifts. While our results suggests that using ResNet-18, a ConvNet from 2015, remains effective with the presence of many models proposed after 2020 and/or based on attention, there is no theoretical guarantee regarding its robustness under major paradigm shifts in the future. 
If that is the case, 
the choice of baseline features might need to be updated accordingly upon new paradigms, which can obviously introduce additional computational overhead.

\subsection{Choosing Sparsity Level in Task Embedding}
\label{sec:eval_choice_L1reg}

In Section \ref{sec:query}, we introduce a scalar hyper-parameter, the sparsity level $\gamma$, to control the strength of sparsity regularization $\gamma \|v\|_1$ when defining the embedding of the downstream task. Here we will present a rule of thumb that we use for choosing $\gamma$ empirically in our experiments.

Intuitively, the sparsity regularization works by penalizing the use of any feature and therefore only features that are critical enough for the downstream task will be utilized. As the sparsity level $\gamma$ increases, the subsets of features preserved will also be smaller.
Informally, to determine the set of features necessary for the downstream task, one can keep increasing the sparsity level $\gamma$ until the downstream performance starts to drop. In Figure \ref{fig:choice_of_L1reg_tmp}, we include downstream accuracy of the baseline features corresponding to varying sparsity level $\gamma$ to present a rule of thumb for deciding the values of sparsity level: simply using the smallest $\gamma$ with at least 3\% accuracy drop from the converged accuracy (i.e. the eventual accuracy when the sparsity level keeps decreasing). More results can be found in Figure \ref{fig:choice_of_L1reg} in Appendix. This single rule is applied to all experiments and it works well as previously presented in Section \ref{sec:eval_performance}.

\section{Conclusion}

We define \textbf{isolated model embedding}, a family of model selection schemes where the update complexity is O(1) and the selection consists of a single sweep over $m$ vectors plus O(1) model operations, both with respect to the number of candidate models $m$.
Isolated model embedding also implies several other desirable properties for applications.
We present \textbf{Standardized Embedder}, an empirical realization of isolated model embedding. Our experiments with 100 vision models support its effectiveness and highlight isolated model embedding as a promising direction towards model selection that is fundamentally (more) scalable. While our experiments focus on visual classifications, the concept of isolated model embedding is modality- and task-agnostic. Thus a natural future direction will be to extend this approach to encompass a broader range of data types and tasks.

\section{Acknowledgement}
Wenxiao Wang would like to thank Samyadeep Basu from University of Maryland for relevant discussions prior to Wenxiao's internship at Sony AI.

\bibliography{ref}
\bibliographystyle{plainnat}


\appendix

\newpage
\section{Full List of Candidate Models Used in The Experiments}
\label{ap:candidates}

\begin{table}[h!]
\caption{A full list of the 100 pre-trained models that are used as the candidate models in the experiments.}
\label{tab:candidates}
\vskip 0.1in
\begin{center}
\resizebox{\linewidth}{!}{
\begin{tabular}{llll||llll}
index & name (used by the corresponding source)                                         & category    & source      & index  & name (used by the corresponding source)                                                    & category                  & source      \\
\hline
1  & ResNet18\_Weights.IMAGENET1K\_V1              & ConvNet     & torchvision & 51  & ShuffleNet\_V2\_X1\_0\_Weights.IMAGENET1K\_V1           & ConvNet                   & torchvision \\
2  & EfficientNet\_B0\_Weights.IMAGENET1K\_V1      & ConvNet     & torchvision & 52  & ShuffleNet\_V2\_X1\_5\_Weights.IMAGENET1K\_V1           & ConvNet                   & torchvision \\
3  & GoogLeNet\_Weights.IMAGENET1K\_V1             & ConvNet     & torchvision & 53  & ShuffleNet\_V2\_X2\_0\_Weights.IMAGENET1K\_V1           & ConvNet                   & torchvision \\
4  & Swin\_T\_Weights.IMAGENET1K\_V1               & Transformer & torchvision & 54  & Swin\_V2\_T\_Weights.IMAGENET1K\_V1                     & Transformer               & torchvision \\
5  & MobileNet\_V3\_Large\_Weights.IMAGENET1K\_V1  & ConvNet     & torchvision & 55  & ViT\_B\_32\_Weights.IMAGENET1K\_V1                      & Transformer               & torchvision \\
6  & MobileNet\_V3\_Large\_Weights.IMAGENET1K\_V2  & ConvNet     & torchvision & 56  & ViT\_B\_16\_Weights.IMAGENET1K\_V1                      & Transformer               & torchvision \\
7  & MobileNet\_V3\_Small\_Weights.IMAGENET1K\_V1  & ConvNet     & torchvision & 57  & ViT\_B\_16\_Weights.IMAGENET1K\_SWAG\_LINEAR\_V1        & Transformer               & torchvision \\
8  & MNASNet0\_5\_Weights.IMAGENET1K\_V1           & ConvNet     & torchvision & 58  & Wide\_ResNet50\_2\_Weights.IMAGENET1K\_V1               & ConvNet                   & torchvision \\
9  & ShuffleNet\_V2\_X0\_5\_Weights.IMAGENET1K\_V1 & ConvNet     & torchvision & 59  & Wide\_ResNet50\_2\_Weights.IMAGENET1K\_V2               & ConvNet                   & torchvision \\
10 & AlexNet\_Weights.IMAGENET1K\_V1               & ConvNet     & torchvision & 60  & mobileone\_s0                                      & ConvNet                   & timm        \\
11 & ConvNeXt\_Tiny\_Weights.IMAGENET1K\_V1        & ConvNet     & torchvision & 61  & mobileone\_s1                                      & ConvNet                   & timm        \\
12 & ConvNeXt\_Small\_Weights.IMAGENET1K\_V1       & ConvNet     & torchvision & 62  & mobileone\_s2                                      & ConvNet                   & timm        \\
13 & DenseNet121\_Weights.IMAGENET1K\_V1           & ConvNet     & torchvision & 63  & mobileone\_s3                                      & ConvNet                   & timm        \\
14 & DenseNet161\_Weights.IMAGENET1K\_V1           & ConvNet     & torchvision & 64  & mobileone\_s4                                      & ConvNet                   & timm        \\
15 & DenseNet169\_Weights.IMAGENET1K\_V1           & ConvNet     & torchvision & 65  & inception\_next\_tiny.sail\_in1k                   & ConvNet                   & timm        \\
16 & DenseNet201\_Weights.IMAGENET1K\_V1           & ConvNet     & torchvision & 66  & inception\_next\_small.sail\_in1k                  & ConvNet                   & timm        \\
17 & EfficientNet\_B1\_Weights.IMAGENET1K\_V1      & ConvNet     & torchvision & 67  & inception\_next\_base.sail\_in1k                   & ConvNet                   & timm        \\
18 & EfficientNet\_B2\_Weights.IMAGENET1K\_V1      & ConvNet     & torchvision & 68  & ghostnet\_100.in1k                                 & ConvNet                   & timm        \\
19 & EfficientNet\_B3\_Weights.IMAGENET1K\_V1      & ConvNet     & torchvision & 69  & ghostnetv2\_100.in1k                               & ConvNet                   & timm        \\
20 & EfficientNet\_B4\_Weights.IMAGENET1K\_V1      & ConvNet     & torchvision & 70  & ghostnetv2\_130.in1k                               & ConvNet                   & timm        \\
21 & EfficientNet\_V2\_S\_Weights.IMAGENET1K\_V1   & ConvNet     & torchvision & 71  & ghostnetv2\_160.in1k                               & ConvNet                   & timm        \\
22 & Inception\_V3\_Weights.IMAGENET1K\_V1         & ConvNet     & torchvision & 72  & repghostnet\_050.in1k                              & ConvNet                   & timm        \\
23 & MNASNet0\_75\_Weights.IMAGENET1K\_V1          & ConvNet     & torchvision & 73  & repghostnet\_058.in1k                              & ConvNet                   & timm        \\
24 & MNASNet1\_0\_Weights.IMAGENET1K\_V1           & ConvNet     & torchvision & 74  & repghostnet\_080.in1k                              & ConvNet                   & timm        \\
25 & MNASNet1\_3\_Weights.IMAGENET1K\_V1           & ConvNet     & torchvision & 75  & repghostnet\_100.in1k                              & ConvNet                   & timm        \\
26 & MobileNet\_V2\_Weights.IMAGENET1K\_V1         & ConvNet     & torchvision & 76  & efficientvit\_b0.r224\_in1k                        & Transformer               & timm        \\
27 & MobileNet\_V2\_Weights.IMAGENET1K\_V2         & ConvNet     & torchvision & 77  & efficientvit\_b1.r224\_in1k                        & Transformer               & timm        \\
28 & RegNet\_X\_1\_6GF\_Weights.IMAGENET1K\_V1     & ConvNet     & torchvision & 78  & efficientvit\_b2.r224\_in1k                        & Transformer               & timm        \\
29 & RegNet\_X\_1\_6GF\_Weights.IMAGENET1K\_V2     & ConvNet     & torchvision & 79  & efficientvit\_b3.r224\_in1k                        & Transformer               & timm        \\
30 & RegNet\_X\_3\_2GF\_Weights.IMAGENET1K\_V1     & ConvNet     & torchvision & 80  & efficientvit\_m0.r224\_in1k                        & Transformer               & timm        \\
31 & RegNet\_X\_3\_2GF\_Weights.IMAGENET1K\_V2     & ConvNet     & torchvision & 81  & efficientvit\_m1.r224\_in1k                        & Transformer               & timm        \\
32 & RegNet\_X\_400MF\_Weights.IMAGENET1K\_V1      & ConvNet     & torchvision & 82  & efficientvit\_m2.r224\_in1k                        & Transformer               & timm        \\
33 & RegNet\_X\_400MF\_Weights.IMAGENET1K\_V2      & ConvNet     & torchvision & 83  & efficientvit\_m3.r224\_in1k                        & Transformer               & timm        \\
34 & RegNet\_X\_800MF\_Weights.IMAGENET1K\_V1      & ConvNet     & torchvision & 84  & efficientvit\_m4.r224\_in1k                        & Transformer               & timm        \\
35 & RegNet\_X\_800MF\_Weights.IMAGENET1K\_V2      & ConvNet     & torchvision & 85  & efficientvit\_m5.r224\_in1k                        & Transformer               & timm        \\
36 & RegNet\_Y\_1\_6GF\_Weights.IMAGENET1K\_V1     & ConvNet     & torchvision & 86  & coatnet\_nano\_rw\_224.sw\_in1k                    & Hybrid (Conv + Attention) & timm        \\
37 & RegNet\_Y\_1\_6GF\_Weights.IMAGENET1K\_V2     & ConvNet     & torchvision & 87  & coatnext\_nano\_rw\_224.sw\_in1k                   & Hybrid (Conv + Attention) & timm        \\
38 & RegNet\_Y\_3\_2GF\_Weights.IMAGENET1K\_V1     & ConvNet     & torchvision & 88  & seresnext101\_32x4d.gluon\_in1k                    & ConvNet                   & timm        \\
39 & RegNet\_Y\_3\_2GF\_Weights.IMAGENET1K\_V2     & ConvNet     & torchvision & 89  & vit\_tiny\_r\_s16\_p8\_224.augreg\_in21k           & Transformer               & timm        \\
40 & RegNet\_Y\_400MF\_Weights.IMAGENET1K\_V1      & ConvNet     & torchvision & 90  & vit\_small\_r26\_s32\_224.augreg\_in21k            & Transformer               & timm        \\
41 & RegNet\_Y\_400MF\_Weights.IMAGENET1K\_V2      & ConvNet     & torchvision & 91  & vit\_tiny\_r\_s16\_p8\_224.augreg\_in21k\_ft\_in1k & Transformer               & timm        \\
42 & RegNet\_Y\_800MF\_Weights.IMAGENET1K\_V1      & ConvNet     & torchvision & 92  & vit\_small\_r26\_s32\_224.augreg\_in21k\_ft\_in1k  & Transformer               & timm        \\
43 & RegNet\_Y\_800MF\_Weights.IMAGENET1K\_V2      & ConvNet     & torchvision & 93  & hrnet\_w18\_small.gluon\_in1k                      & ConvNet                   & timm        \\
44 & ResNeXt50\_32X4D\_Weights.IMAGENET1K\_V1      & ConvNet     & torchvision & 94  & hrnet\_w18\_small\_v2.gluon\_in1k                  & ConvNet                   & timm        \\
45 & ResNeXt50\_32X4D\_Weights.IMAGENET1K\_V2      & ConvNet     & torchvision & 95  & vit\_small\_patch16\_224.dino                      & Transformer               & timm        \\
46 & ResNet101\_Weights.IMAGENET1K\_V1             & ConvNet     & torchvision & 96  & vit\_base\_patch16\_224.mae                        & Transformer               & timm        \\
47 & ResNet101\_Weights.IMAGENET1K\_V2             & ConvNet     & torchvision & 97  & maxvit\_tiny\_tf\_224.in1k                         & Hybrid (Conv + Attention) & timm        \\
48 & ResNet50\_Weights.IMAGENET1K\_V1              & ConvNet     & torchvision & 98  & maxvit\_tiny\_rw\_224.sw\_in1k                     & Hybrid (Conv + Attention) & timm        \\
49 & ResNet50\_Weights.IMAGENET1K\_V2              & ConvNet     & torchvision & 99  & vit\_base\_patch32\_224.sam\_in1k                  & Transformer               & timm        \\
50 & ResNet34\_Weights.IMAGENET1K\_V1              & ConvNet     & torchvision & 100 & vit\_base\_patch32\_clip\_224.openai\_ft\_in1k     & Transformer               & timm       
\end{tabular}
}
\end{center}
\vskip -0.1in
\end{table}

In Table \ref{tab:candidates}, we include the full list of pre-trained models that are used in our evaluations. We include as follows relevant references and corresponding model indices in Table \ref{tab:candidates} (note that some pre-trained models correspond to multiple references):
\begin{itemize}
    \item ResNet \citep{ResNet}: 1, 46, 47, 48, 49, 50, 88;
    \item EfficientNet/EfficientNetV2 \citep{EfficientNet, EfficientNetV2}: 2, 17, 18, 19, 20, 21;
    \item GoogLeNet \citep{GoogLeNet}: 3;
    \item Swin Transformer/Swin Transformer V2 \citep{SwinT, SwinT2}: 4, 54;
    \item MobileNet V2/V3 \citep{mobilenetv2, mobilenetv3}: 5, 6, 7, 26, 27;
    \item MNASNet \citep{MNASNet}: 8, 23, 24, 25;
    \item ShuffleNet V2 \citep{ShuffleNet}: 9, 51, 52, 53;
    \item AlexNet \citep{AlexNet}: 10;
    \item ConvNeXt \citep{convnext}: 11, 12, 87;
    \item DenseNet \citep{DenseNet}: 13, 14, 15, 16;
    \item Inception V3 \citep{inceptionv3}: 22;
    \item RegNet \citep{regnet}: 28, 29, 30, 31, 32, 33, 34, 35, 36, 37, 38, 39, 40, 41, 42, 43;
    \item ResNeXt \citep{resnext}: 44, 45, 88;
    \item Vision Transformer \citep{vit}: 55, 56, 57, 89, 90, 91, 92, 95, 96, 99, 100;
    \item Wide ResNet \citep{wide_resnet}: 58, 59;
    \item MobileOne \citep{mobileone}: 60, 61, 62, 63, 64;
    \item InceptionNeXt \citep{inceptionnext}: 65, 66, 67;
    \item GhostNet/GhostNetV2 \citep{ghostnet, ghostnetv2}: 68, 69, 70, 71;
    \item RepGhostNet \citep{repghost}: 72, 73, 74, 75;
    \item EfficientViT (MIT) \citep{efficientvit_mit}: 76, 77, 78, 79;
    \item EfficientViT (MSRA) \citep{efficientvit_msra}: 80, 81, 82, 83, 84, 85;
    \item CoAtNet \citep{coatnet}: 86, 87;
    \item Squeeze-and-Excitation \citep{SE}: 88;
    \item Bag-of-Tricks \citep{bagoftrick}: 88;
    \item AugReg \citep{augreg}: 89, 90, 91, 92;
    \item HRNet \citep{hrnet}: 93, 94;
    \item DINO \citep{DINO}: 95;
    \item Masked Autoencoder \citep{MAE}: 96;
    \item MaxViT \citep{maxvit}: 97, 98;
    \item Sharpness-aware minimizer for ViT \citep{samvit}: 99;
    \item Reproducible scaling laws \citep{repro_scaling_law}: 100;
    \item CLIP \citep{clip}: 100.
\end{itemize}

\newpage
\section{Figures of Empirical Evaluations}

\begin{figure}[H]
\vskip 0.1in
\begin{center}
\subfigure[downstream: CIFAR-10\newline
baseline features: ResNet-18]{
\includegraphics[width=0.3\linewidth]{figures/cifar10_resnet18.png}
}
\hfill
\subfigure[downstream: CIFAR-100\newline
baseline features: ResNet-18]{
\includegraphics[width=0.3\linewidth]{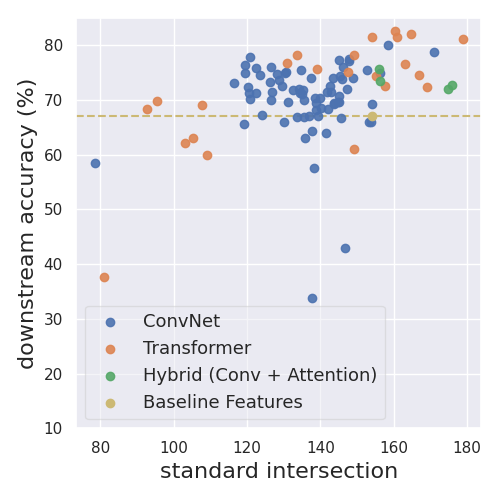}
}
\hfill
\subfigure[downstream: STL-10\newline
baseline features: ResNet-18]{
\includegraphics[width=0.3\linewidth]{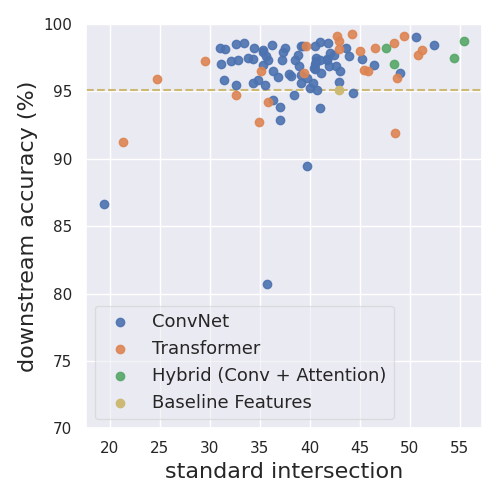}
}

\subfigure[downstream: CIFAR-10\newline
baseline features: Swin-T (tiny)]{
\includegraphics[width=0.3\linewidth]{figures/cifar10_swin_t.png}
}\hfill
\subfigure[downstream: CIFAR-100\newline
baseline features: Swin-T (tiny)]{
\includegraphics[width=0.3\linewidth]{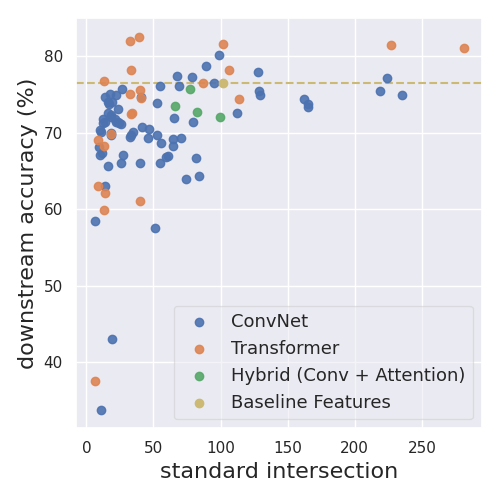}
}\hfill
\subfigure[downstream: STL-10\newline
baseline features: Swin-T (tiny)]{
\includegraphics[width=0.3\linewidth]{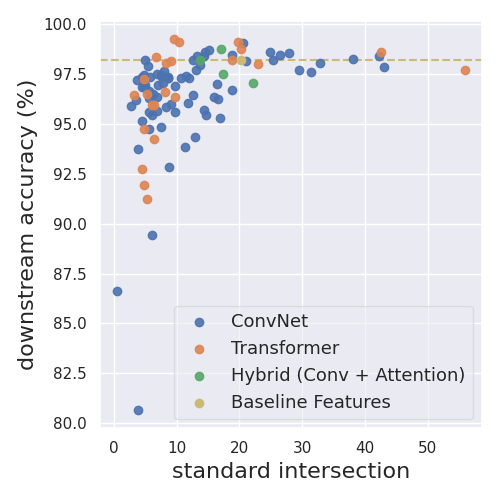}
}
\caption{Downstream accuracy (i.e. the ground truth) v.s. the cardinality of standard intersections (i.e. the selection metric) when using 4k steps per candidate. The downstream accuracy of the baseline features are highlighted with the \textcolor{olive}{dashed line}. When a public model is only suboptimal, using it as baseline features for \name can still locate more competitive models.}
\label{fig:acc_vs_si}
\end{center}
\vskip -0.1in
\end{figure}
\null
\vfill
\newpage

\begin{figure}[H]
\vskip 0.1in
\begin{center}
\subfigure[downstream: CIFAR-10\newline
baseline features: ResNet-18]{
\includegraphics[width=0.3\linewidth]{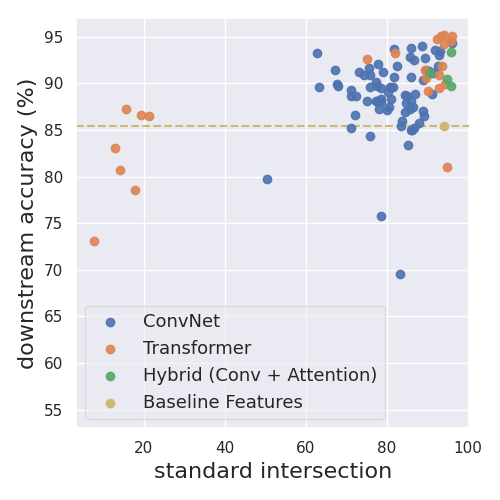}
}\hfill
\subfigure[downstream: CIFAR-100\newline
baseline features: ResNet-18]{
\includegraphics[width=0.3\linewidth]{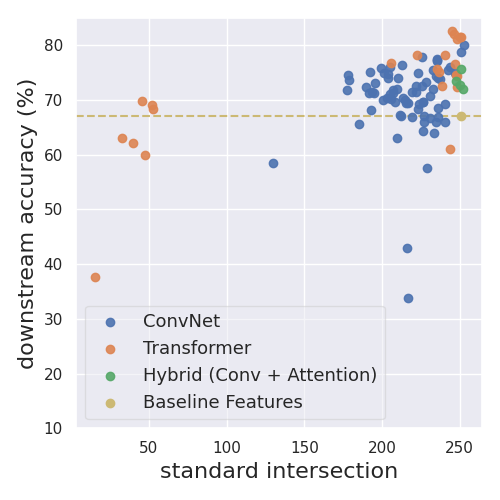}
}\hfill
\subfigure[downstream: STL-10\newline
baseline features: ResNet-18]{
\includegraphics[width=0.3\linewidth]{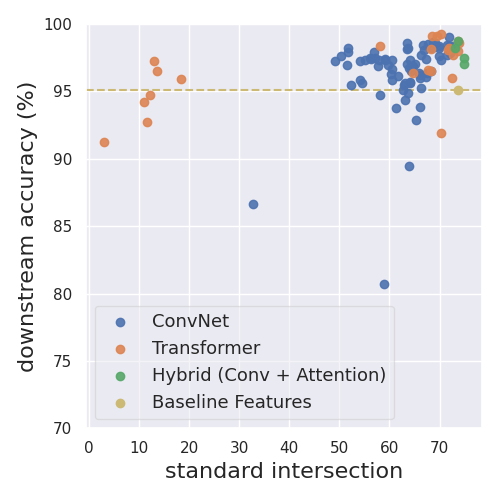}
}

\subfigure[downstream: CIFAR-10\newline
baseline features: Swin-T (tiny)]{
\includegraphics[width=0.3\linewidth]{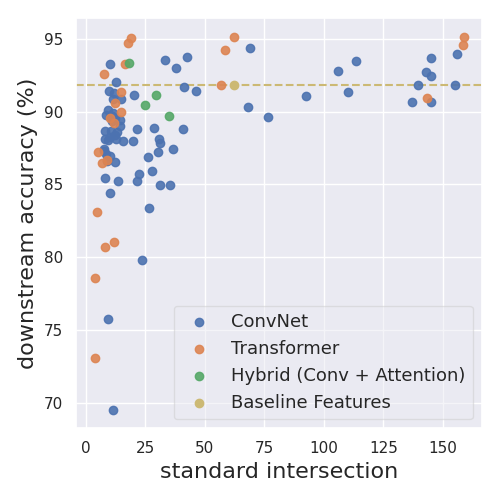}
}\hfill
\subfigure[downstream: CIFAR-100\newline
baseline features: Swin-T (tiny)]{
\includegraphics[width=0.3\linewidth]{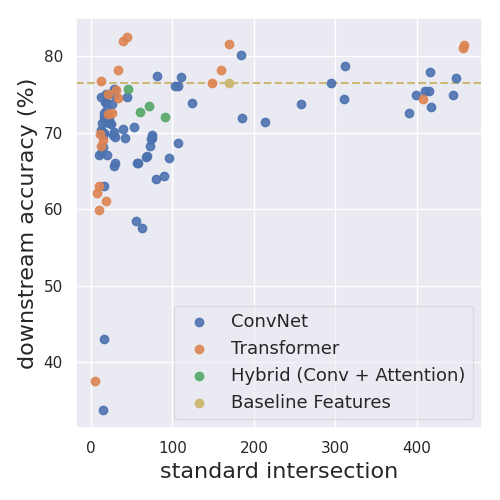}
}\hfill
\subfigure[downstream: STL-10\newline
baseline features: Swin-T (tiny)]{
\includegraphics[width=0.3\linewidth]{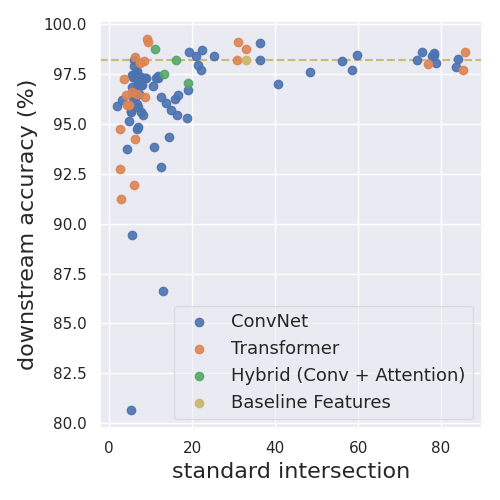}
}
\caption{Downstream accuracy (i.e. the ground truth) v.s. the cardinality of standard intersections (i.e. the selection metric) when using 10k steps per candidate. The downstream accuracy of the baseline features are highlighted with the \textcolor{olive}{dashed line}. When a public model is only suboptimal, using it as baseline features for \name can still locate more competitive models.}
\label{fig:acc_vs_si_10k}
\end{center}
\vskip -0.1in
\end{figure}
\null
\vfill


\newpage
\begin{figure}[tbp!]
\vskip 0.1in
\begin{center}
\subfigure[downstream: CIFAR-10]{
\includegraphics[width=0.3\linewidth]{figures/cmp_CIFAR10_4k.png}
}\hfill
\subfigure[downstream: CIFAR-100]{
\includegraphics[width=0.3\linewidth]{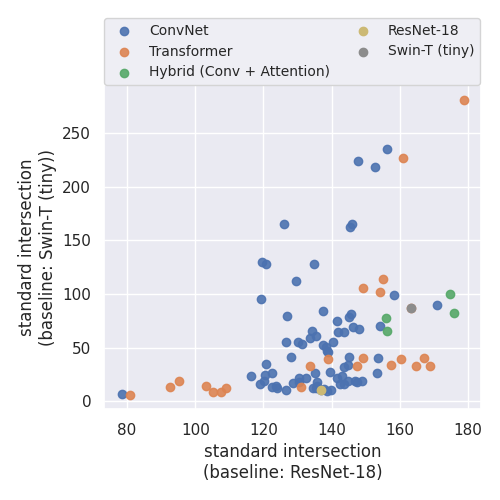}
}\hfill
\subfigure[downstream: STL-10]{
\includegraphics[width=0.3\linewidth]{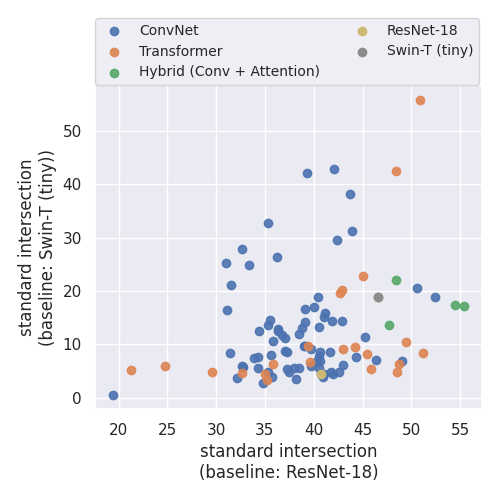}
}
\caption{Comparing the cardinality of standard intersections (i.e. the selection metric) when using different baseline features (ResNet-18 and Swin-T (tiny)) with 4k steps per candidate. The green/orange points in the bottom right suggest using ResNet-18 as baseline features tend to overestimate (some) models with attentions compared to using Swin Transformer (tiny).}
\label{fig:si_vs_si_4k}
\end{center}
\vskip -0.1in
\end{figure}

\begin{figure}[H]
\vskip 0.1in
\begin{center}
\subfigure[downstream: CIFAR-10]{
\includegraphics[width=0.3\linewidth]{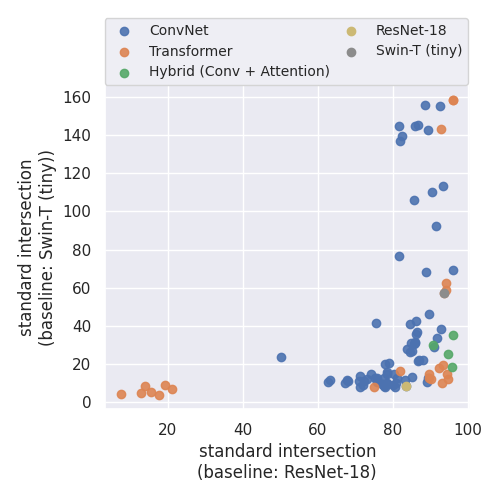}
}\hfill
\subfigure[downstream: CIFAR-100]{
\includegraphics[width=0.3\linewidth]{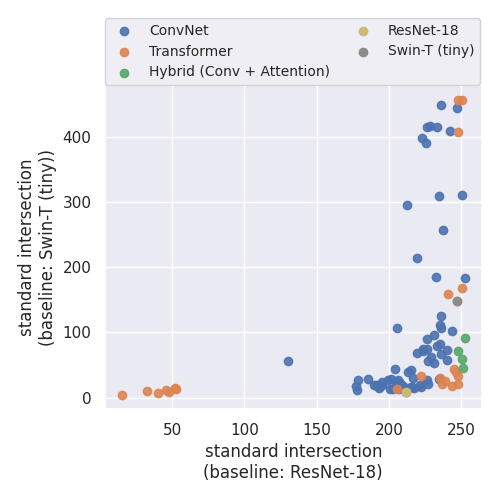}
}\hfill
\subfigure[downstream: STL-10]{
\includegraphics[width=0.3\linewidth]{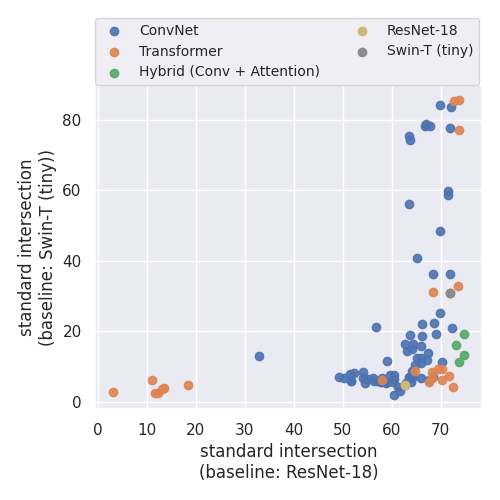}
}
\caption{Comparing the cardinality of standard intersections (i.e. the selection metric) when using different baseline features (ResNet-18 and Swin-T (tiny)) with 10k steps per candidate. The green/orange points in the bottom right suggest using ResNet-18 as baseline features tend to overestimate (some) models with attentions compared to using Swin Transformer (tiny).}
\label{fig:si_vs_si_10k}
\end{center}
\vskip -0.1in
\end{figure}
\null
\vfill
\newpage

\begin{figure}[h!]
\vskip 0.1in
\begin{center}
\subfigure[downstream: CIFAR-10\newline
baseline features: ResNet-18]{
\includegraphics[width=0.3\linewidth]{figures/cifar10_resnet18_gamma.png}
}\hfill
\subfigure[downstream: CIFAR-100\newline
baseline features: ResNet-18]{
\includegraphics[width=0.3\linewidth]{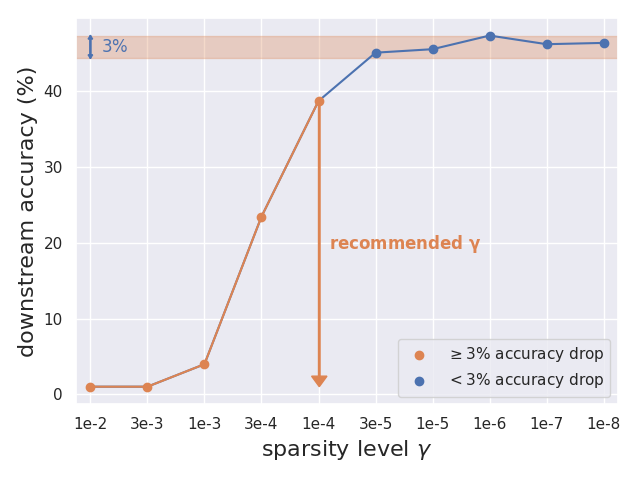}
}\hfill
\subfigure[downstream: STL-10\newline
baseline features: ResNet-18]{
\includegraphics[width=0.3\linewidth]{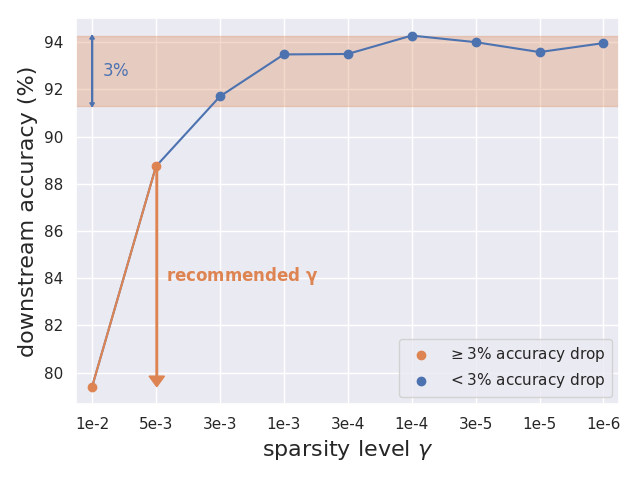}
}

\subfigure[downstream: CIFAR-10\newline
baseline features: Swin-T (tiny)]{
\includegraphics[width=0.3\linewidth]{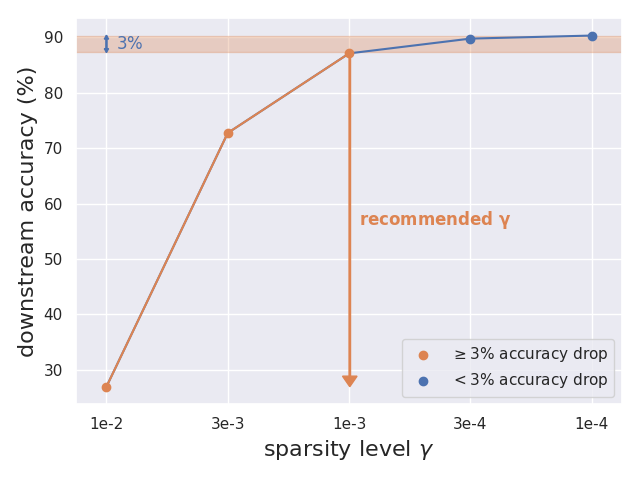}
}\hfill
\subfigure[downstream: CIFAR-100\newline
baseline features: Swin-T (tiny)]{
\includegraphics[width=0.3\linewidth]{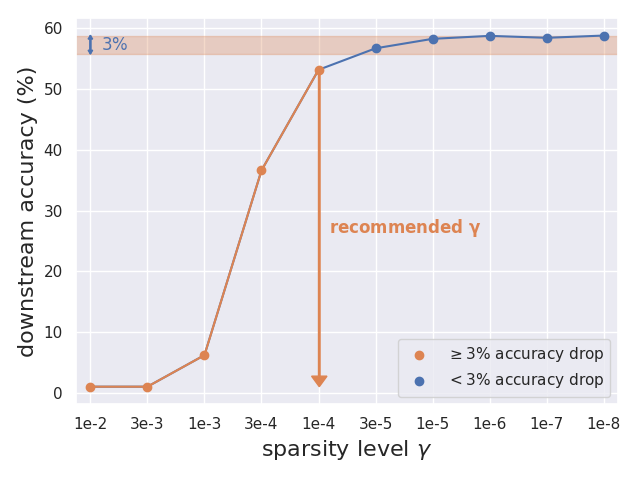}
}\hfill
\subfigure[downstream: STL-10\newline
baseline features: Swin-T (tiny)]{
\includegraphics[width=0.3\linewidth]{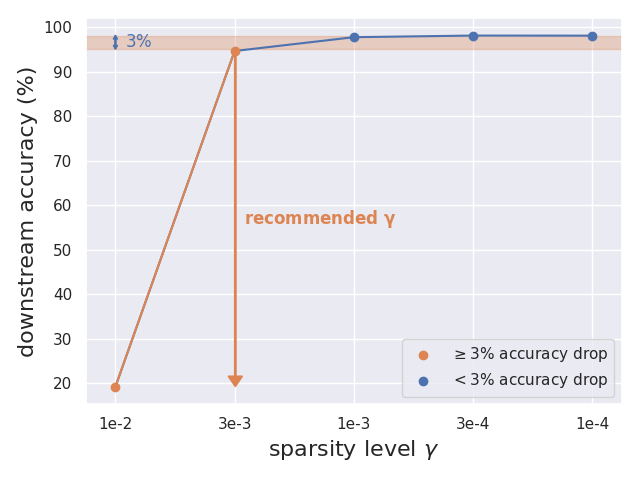}
}
\caption{Downstream accuracy of the baseline features corresponding to varying level of sparsity regularization $\gamma$. A rule of thumb for deciding the value of $\gamma$: using the smallest $\gamma$ with at least 3\% accuracy drop from the converged accuracy.}
\label{fig:choice_of_L1reg}
\end{center}
\vskip -0.1in
\end{figure}

\section{Illustrative Example for Section \ref{sec:preparation}}
\label{appendix:toyexample}
Here is an example to illustrate how to associate every subset of the baseline feature set $B$ with a binary vector from $\{0, 1\}^N$. Assuming now the baseline feature set $B: \mathcal{X} \to \mathbb{R}^N$ contains a total of $N=4$ features, $b_0, b_1, b_2, b_3$, where each of them is a function from $\mathcal{X}$ to $\mathbb{R}$, then there will be a total of $2^4=16$ different subsets of $B$. We can associate each subset with a distinct, 4-dimensional binary vector (i.e. a vector in $\{0,1\}^4$) by using $1$ to indicate the presence of a feature and $0$ to indicate an absence of a feature in the subset. Specifically, $(0, 0, 0, 0)$ will denote the empty subset, $(0, 0, 1, 0)$ will denote $\{b_2\}$ and $(1, 0, 1, 1)$ will denote $\{b_0, b_2, b_3\}$.

\section{Broader Impacts}
\label{appendix: impact}

When selecting models locally, asymptotically fast update and selection indicate considerably improved scalability, which is obviously positive as one can potentially utilize more candidate models. 
Meanwhile, the decentralizability of isolated model embedding naturally relates to a potential application: a decentralized model market with a decentralized model selection system. 

There are, of course, positive impacts from having such a decentralized model market. For individual downstream users, their capabilities can be extended by effectively accessing more candidate models, and their costs for model selection can be potentially reduced through systematic and out-of-the-box selection operations. 
For model owners, their profits can be enlarged by promoting their models to more potential users.

However, while the benefits of such a decentralized market stem from the accelerated distribution of models, the distribution of malicious behaviors, such as backdoors, may also be facilitated in the presence of hostile parties, which is potentially a negative impact and a new research direction.


\end{document}